%% file: videosampling.tex
\documentclass[tog]{acmsiggraph}

\def\BibTeX{{\rm B\kern-.05em{\sc i\kern-.025em b}\kern-.08em
    T\kern-.1667em\lower.7ex\hbox{E}\kern-.125emX}}

\input{includes}

\TOGonlineid{0202}
\TOGvolume{0}
\TOGnumber{0}
\TOGarticleDOI{1111111.2222222}

\title{Sampling Based Scene-Space Video Processing}
\author{Felix Klose$^{12\ast}$ \qquad Oliver Wang$^{1\ast}$ \qquad Jean-Charles Bazin$^1$ \qquad Marcus Magnor$^2$ \qquad Alexander Sorkine-Hornung$^1$ \\
		$^1$ Disney Research Zurich \qquad $^2$ TU Braunschweig}
\pdfauthor{Felix Klose}


\begin{document}

\teaser{
\setlength{\tabcolsep}{1px}
\centering
	\def\myheight{4.6cm}
	\begin{tabular}{*{2}{c@{\hspace{2px}}}}
	\includegraphics[clip,trim=75 0 740 0,height = \myheight]{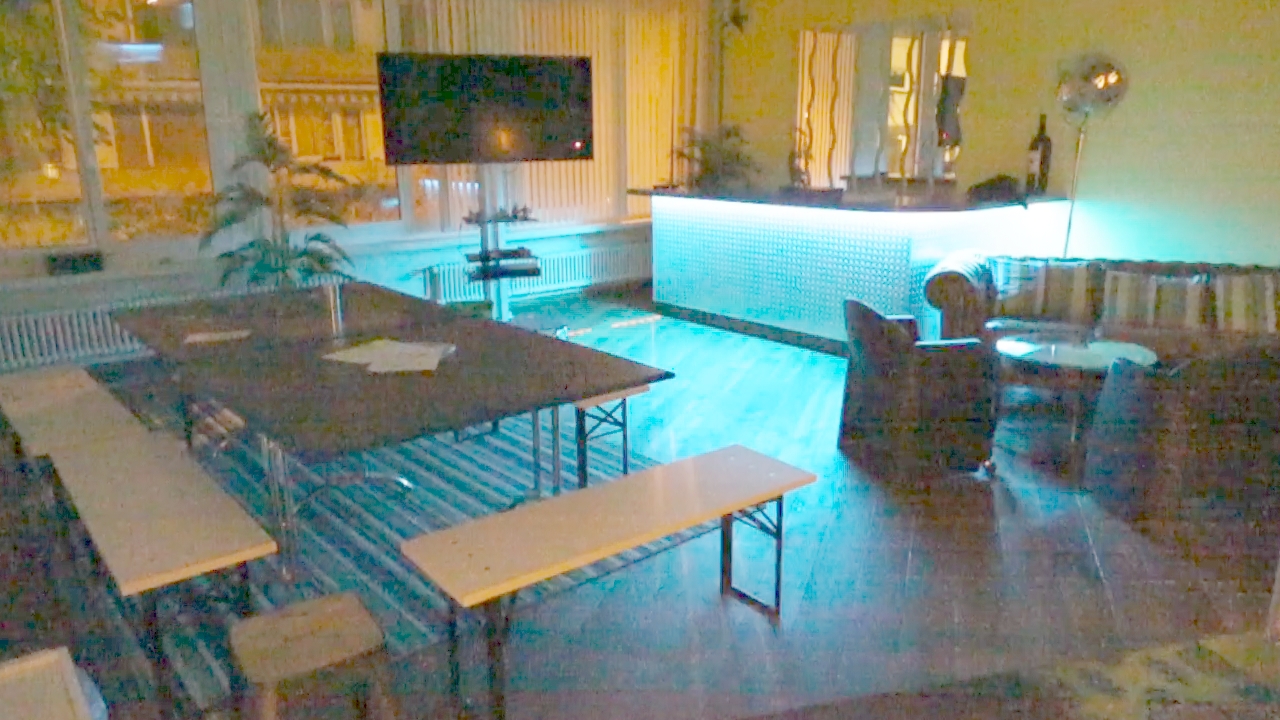} &
	\includegraphics[clip,trim=540 0 275 0,height = \myheight]{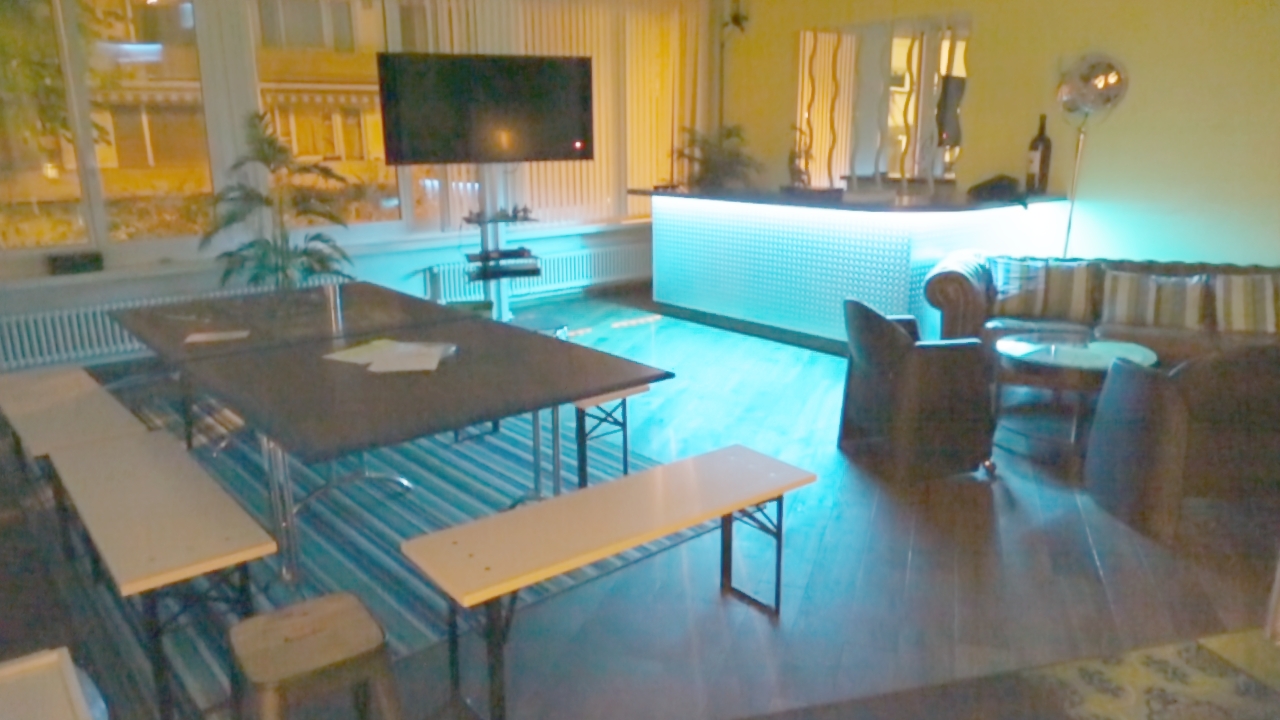} \\
	\multicolumn{2}{c}{Denoising} \\	
	\end{tabular}
	\begin{tabular}{*{1}{c@{\hspace{1px}}}}
	\includegraphics[clip,trim=200 0 80 0,height = \myheight]{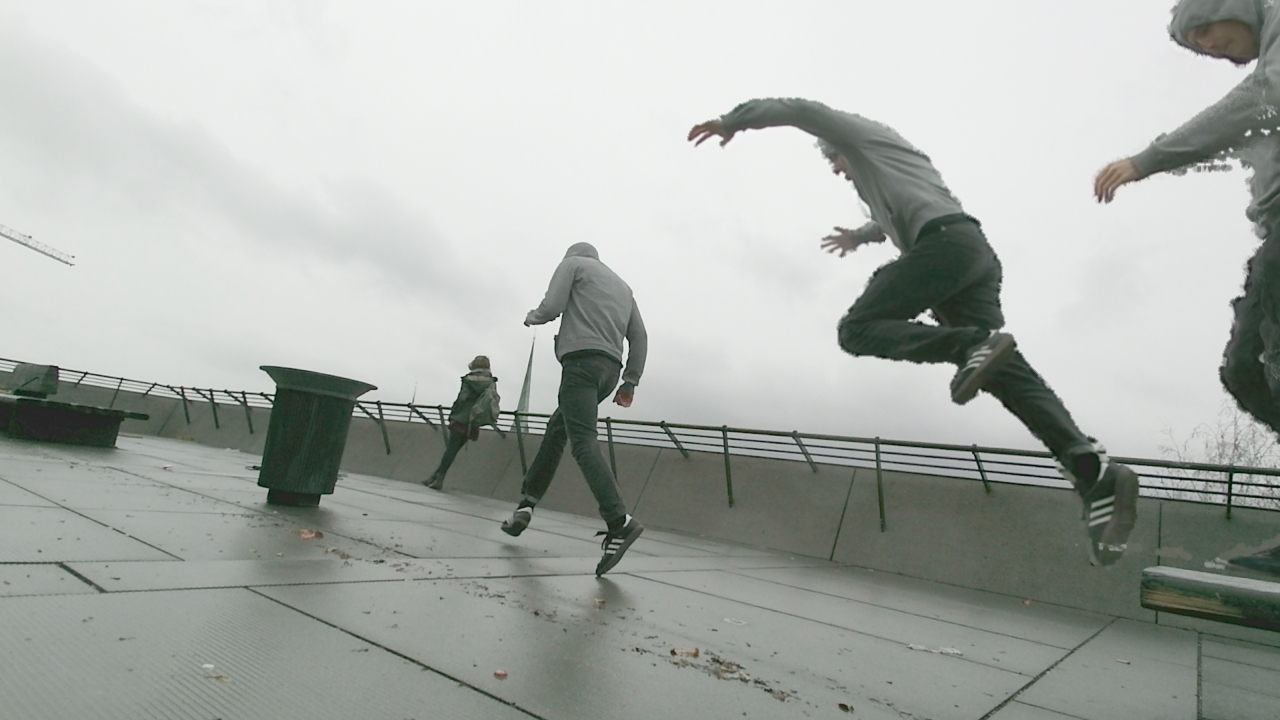} \\
	Action shots \\
	\end{tabular}
	\begin{tabular}{*{1}{c@{\hspace{5px}}}}
	\includegraphics[clip,trim=450 100 250 25,height = \myheight]{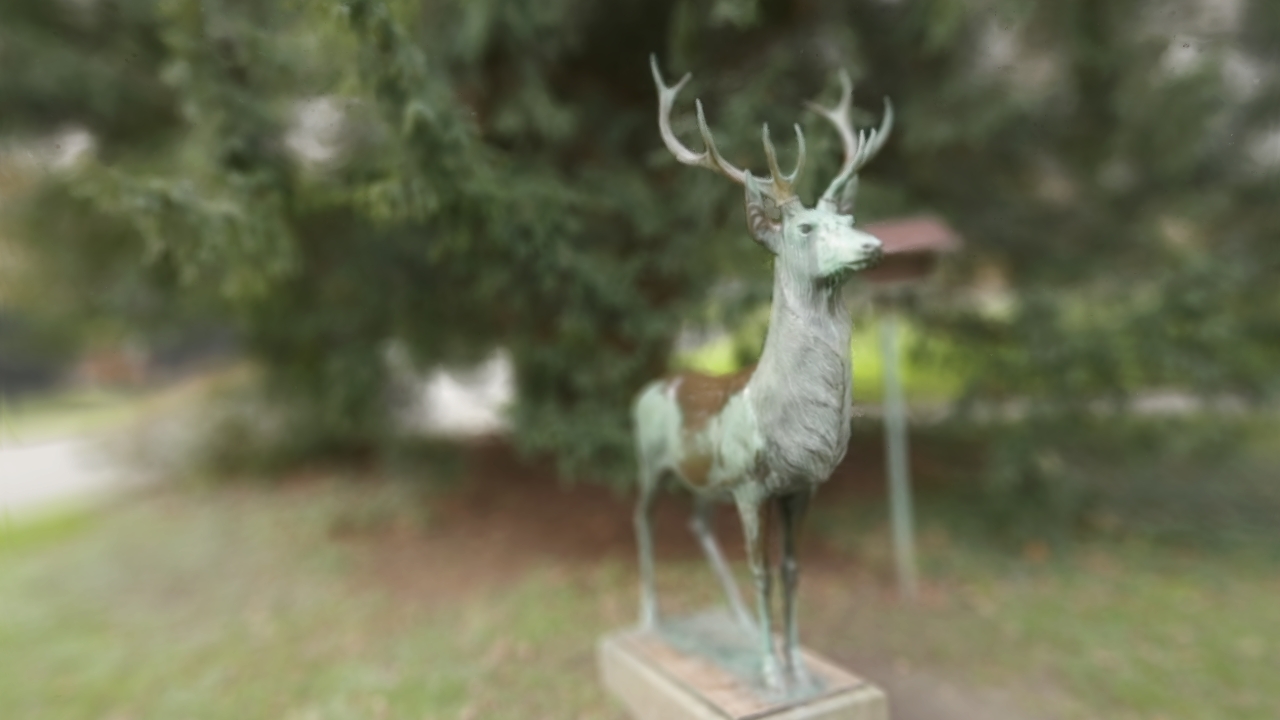} \\
	Virtual aperture \\
	\end{tabular}
    \caption{Single frames from video results created with our sampling based scene-space video processing framework. It enables fundamental video applications such as denoising (left) as well as new artistic results such as action shots (center) and virtual aperture effects (right).  Our approach is robust to unavoidable inaccuracies in 3D information, and can be used on casually recorded, moving video.}
    \label{fig:teaser}
}

\maketitle

\vspace{-1cm}
\begin{abstract}
\input{abstract}

\end{abstract}

\copyrightspace

\begin{CRcatlist}
  \CRcat{I.2.10}{Computer Graphics}{Picture/Image Generation}{Display Algorithms}
  \CRcat{I.3.7}{Artificial Intelligence}{Vision and Scene Understanding}{3D/stereo scene analysis};
\end{CRcatlist}

\keywordlist

\blfootnote{$^\ast$ denotes joint first authorship with equal contribution}

\section{Introduction}
\label{sec:introduction}
\input{introduction}

\section{Related Work}
\label{sec:relatedwork}
\input{relatedwork}

\section{Method Overview}
\label{sec:overview}
\input{method}

\section{Sample Gathering}
\label{sec:sampling}
\input{sampling}

\section{Filtering}
\label{sec:processing}
\input{processing}

\input{applications}

\section{Implementation}
\label{sec:implementation}

\input{implementation}

\section{Evaluation}
\label{sec:evaluation}
\input{evaluation}

\section{Conclusion}
\label{sec:conclusion}
\input{conclusion}

\section*{Acknowledgments}
\final{
We would like to thank Henning Zimmer, Changil Kim, Ya\u{g}{\i}z Aksoy, Cengiz \"{O}ztireli, and Mario Botsch for helpful discussions throughout the project, as well as Sunghyun Cho and Miguel Granados for making their datasets public. Marcus Magnor acknowledges funding from ERC Grant \#256942 ``RealityCG''.
}

\bibliographystyle{acmsiggraph}
\bibliography{videosampling}
\end{document}

%% file: includes.tex
\usepackage[usenames,dvipsnames]{color}
\usepackage{graphicx}
\usepackage{pgfplots}
\usepackage{tikz}
\usetikzlibrary{spy, shapes, arrows, matrix, pgfplots.groupplots}
\usepackage{amsmath}
\usepackage{amssymb}
\usepackage{subfig}
\usepackage{enumitem}
\usepackage{todonotes}
\usepackage{bm}
\usepackage{booktabs}
\usepackage{rotating}
\usepackage{multirow}

\definecolor{myorchid}{RGB}{20,150,30}
\definecolor{superfancycolorname}{RGB}{200,0,0}
\definecolor{myblue}{RGB}{10,10,200}
\definecolor{somenicecolor}{RGB}{0,206,209}
\definecolor{burntorang}{RGB}{100,100,0}

\newcommand{\ow}[1] {\emph{\textcolor{myorchid}{OW: #1}}}

\newcommand{\ignore}[1] {}
\newcommand{\final}[1]{#1}

\newcommand\blfootnote[1]{%
  \begingroup
  \renewcommand\thefootnote{}\footnote{#1}%
  \addtocounter{footnote}{-1}%
  \endgroup
}

%% file: abstract.tex
Many compelling video processing effects can be achieved if per-pixel depth information and 3D camera calibrations are known.
However, the success of such methods is highly dependent on the accuracy of this ``scene-space'' information.
We present a novel, sampling-based framework for processing video that enables high-quality scene-space video effects in the presence of inevitable errors in depth and camera pose estimation.
Instead of trying to improve the explicit 3D scene representation, the key idea of our method is to exploit the high redundancy of approximate scene information that arises due to most scene points being visible multiple times across many frames of video.
Based on this observation, we propose a novel pixel gathering and filtering approach. %
The gathering step is general and collects pixel samples in scene-space, while the filtering step is application-specific and computes a desired output video from the gathered sample sets.
Our approach is easily parallelizable and has been implemented on GPU, allowing us to take full advantage of large volumes of video data and facilitating practical runtimes on HD video using a standard desktop computer.
Our generic scene-space formulation is able to comprehensively describe a multitude of video processing applications such as denoising, deblurring, super resolution, object removal, computational shutter functions, and other scene-space camera effects.
We present results for various casually captured, hand-held, moving, compressed, monocular videos depicting challenging scenes recorded in uncontrolled environments.

%% file: introduction.tex
Scene-space video processing, where pixels are processed according to their 3D positions, has many advantages over traditional image-space processing.
For example, handling camera motion, occlusions, and temporal continuity entirely in 2D image-space can in general be very challenging, while dealing with these issues in scene-space is simple. 
As scene-space information, e.g., in the form of depth maps and camera pose parameters, becomes more and more widely available due to advances in tools and mass market hardware devices (such as portable light field cameras, depth-enabled smart phones~\cite{ProjectTango} and RGBD cameras), techniques that leverage depth information will play an important role in future video processing approaches.

Previous work has shown that accurate scene-space information makes fundamental video processing problems simple, automatic, and robust~\cite{DBLP:conf/rt/BhatZSAACCK07,Zhang_refilming_tvcg09} and even enables the creation of new, compelling video effects~\cite{KholgadeSES14,DBLP:journals/tog/Szeliski14}.
However, the visual output quality of such methods is highly dependent on the quality of the available scene-space information.
Despite considerable advances in 3D reconstruction over the last decades, computing exact 3D information for arbitrary video recorded under uncontrolled conditions remains an elusive goal (and will likely remain so for the foreseeable future, due to inherent ambiguities in these tasks).

We propose an alternative, general-purpose framework that facilitates robust scene-space video processing in the presence of incorrect 3D information by exploiting the high degree of redundancy in video that arises from multiple observations of the same scene point over time.
Our approach takes as input a \emph{casually acquired}, hand-held, moving, and uncalibrated video, possibly altered by compression artifacts, along with potentially incorrect depth maps and camera pose information, both of which are either derived from the input footage itself or acquired using other sensors.
Our work on sample based scene-space processing is inspired by recent advances in edge-aware filtering and consists of the following two steps.
First, for each output pixel an efficient, general-purpose gathering step collects a large set of input video pixels that potentially represent alternative observations of the same scene point.
In the second step, an application-specific filtering operation efficiently reduces the contribution of outliers in the sample set and computes the output pixel color as a weighted combination of the gathered samples.

This sampling-based framework allows us to formulate a wide range of video effects as straightforward, transparent filtering operations in scene space, operating directly on noisy scene-space data without the need for accurate reconstruction or refinement techniques. 
We demonstrate a wide range of practical applications, from fundamental video processing tasks such as denoising, deblurring, and super resolution to advanced video effects such as action shots, virtual apertures, object removal, and computational shutter effects.
We show results computed from different sources of depth information including standard multi-view stereo reconstruction and \final{one application that uses} actively acquired depth from a Kinect.
Because our approach is robust to spurious depth values, we can use simple and fast local depth estimation and deal with the resulting depth noise in the filtering step.
While our approach assumes the existence of a consistent scene-space representation over a number of frames, its robustness to outliers allows us to handle some degree of dynamic content, and we show a variety of results on scenes with moving objects.

The entire framework can be easily parallelized.
We describe a GPU implementation that is capable of gathering billions of samples on-the-fly without the need for explicit storage in a dedicated data structure or quantization of scene-space, achieving practical runtime performance for real-world HD video footage on a desktop computer.

The key idea is that our framework relies on \textbf{simple, fast} and \textbf{transparent} algorithms that achieve high-quality results from inaccurate 3D information by exploiting scene structure redundancy across multiple video frames.

%% file: relatedwork.tex
Our algorithm applies to a wide range of applications in image and video processing. 
In this section, we therefore briefly review related work on a general level and discuss more specific related research in the context of the respective applications in Sec.~\ref{sec:processing}.

\paragraph{Pre-processing}
Our approach leverages recent advances in 3D reconstruction techniques.
Among these are structure-from-motion~(SFM) methods that simultaneously recover scene points and camera poses~\cite{KolevKBC09,FurukawaP10}, and real-time variants like simultaneous localization and mapping~(SLAM)~\cite{NewcombeD10,TanskanenKMCSP13}.
In all our experiments, we estimate camera pose parameters automatically using commonly available commercial tools (NUKEX 9.0v5).
In addition, our scene-space framework requires per-pixel depth needed to project each video pixel into scene-space.
Depth from monocular video is customarily computed using stereo methods employed on pairs of subsequent video frames~\cite{DBLP:journals/ijcv/ScharsteinS02,DBLP:conf/cvpr/SeitzCDSS06}.
Other techniques for generating spatio-temporally consistent depth maps from video sequences rely on segmentation and bundle optimization~\cite{ZhangJWB09} or filtering~\cite{DBLP:journals/tog/LangWASG12}.
Such methods involve complex, computationally expensive global optimization routines to handle inherent ambiguities and/or to enforce smoothness.
In contrast, for our framework we found it completely sufficient to compute dense depth using a much simpler, data-term only depth estimation algorithm, or to use raw depth data from sensors like the Kinect, since our sample collection and filtering step robustly eliminates outliers during rendering.

\paragraph{Depth-image based rendering}
Scene-space information is frequently used in image-based rendering methods to synthesize views of recorded scenes from arbitrary perspectives.
Most commonly, depth-image based rendering (DIBR) methods work by projecting pixels into virtual camera views using per-pixel depth information~\cite{DBLP:journals/tog/ZitnickKUWS04}.
Recent extensions use additional image-based correspondences to mask depth projection errors~\cite{DBLP:journals/tcsv/LipskiKM14}.
For a survey of depth image-based rendering approaches, we refer the reader to~\cite{ShumIBRbook}.
In contrast to the goals of image-based rendering, our work does \emph{not} aim to synthesize novel views from virtual camera positions.
We are instead focused on processing the recorded video frames themselves.
This key distinction allows us to achieve high visual fidelity and realism by exploiting the original frames as a prior.

\paragraph{Point-based methods}
Similar to our approach, point-based rendering techniques deal with large numbers of unstructured points.
These methods work by ``splatting'' unstructured, oriented 3D point clouds into a virtual camera view for displaying complex scenes~\cite{DBLP:conf/siggraph/ZwickerPBG01}.
Similar to our approach, these methods accumulate surface samples in screen space.
In order to cope with noisy outlier points, it is common in point-based rendering to resample and filter the 3D point cloud, e.g., by moving least squares~\cite{AlexaBCFLS03,OtireliGG09,KusterBODMPG14}. %
These methods focus on rendering predominantly correct 3D points corrupted by spurious noise (often acquired by a laser scanner), and obtain robustness by assuming that the underlying geometry is a locally smooth manifold.
Epipolar constraints have been shown to improve rendering quality for transitions between views, despite some inaccurate depth estimates~\cite{GoeseleAFHKD10}.
Our framework, in contrast, is designed to handle cases where the vast majority of samples are outliers that arise from incorrect 3D estimates.
Our approach does not assume a spatially smooth model, but relies on redundancy from very large sample sets (on the order of billions of samples or more) and a bilateral weighting scheme to remove the contribution of outliers.
Our target application is general video processing as opposed to 3D rendering.

\paragraph{Filtering}
Our work is inspired in part by the recent successes of ``edge-aware'' filtering methods in image and video processing~\cite{paris2007gentle}.
Such approaches are often used for image processing where an image-space patch is filtered by a weighted combination of pixels based on a multivariate normal distribution centered around the input pixel.
This simple idea has been shown to be successful in handling a large degree of outliers and has been used in a wide range of applications including tonemapping and style transfer~\cite{DBLP:journals/tog/AubryPHKD14}, upsampling~\cite{DBLP:journals/tog/KopfCLU07}, colorization~\cite{DBLP:journals/tog/GastalO11}, and to approximate global regularizers~\cite{DBLP:journals/tog/LangWASG12}.
\final{
Zhang et al.~\shortcite{DBLP:conf/cvpr/ZhangVJN09} propose a method that leverages multiple view geometry to find image patches for denoising.
This approach marginalizes over depth values, and is therefore somewhat robust to bad depth estimations, however it works only for single images. 
}
Our filtering step resembles these methods, but instead of deriving weights from image patches we filter samples collected from \emph{scene-space frustums} in a high-dimensional sample space.

\paragraph{Depth-aware video enhancement}
Prior approaches have used depth to achieve various video effects such as stylization and relighting~\cite{DBLP:journals/cgf/RichardtSDST12}, manipulating still images by registering stock 3D models to image content~\cite{KholgadeSES14}, or computing visually consistent hyperlapses~\cite{DBLP:journals/tog/Szeliski14}.
Other methods have proposed registering high-resolution stills to improve the quality of video effects such as HDR, superresolution and object removal on static~\cite{DBLP:conf/rt/BhatZSAACCK07} as well as dynamic~\cite{gupta:2009} scenes.
Zhang et al.~\shortcite{Zhang_refilming_tvcg09} generate transparency, bullet time, and depth-of-field effects using high-quality, computationally expensive depth maps.
While these approaches generate compelling results, they rely on accurate scene information.
Consequently, the quality of their result is directly affected by any errors in the estimated 3D scene correspondences.
Our approach, in contrast, presents a new way of operating in scene-space that is robust to noisy depth.

Related to our work, Sunkavalli et al.~\shortcite{DBLP:journals/tvcg/SunkavalliJKCP12} generate novel still images from short videos by aligning video frames and computing importance-based pixel weights on the resulting image stack.
They show applications such as super resolution, blur and noise reduction, as well as visual summaries (action shots).
While this approach creates compelling snapshots, it operates entirely in image-space and does not straightforwardly extend to video.
By regarding the task in scene-space instead of image-space, our framework may be considered a generalization of this earlier work that is also applicable to videos recorded by a moving camera.

%% file: method.tex
\begin{figure}[t]
	\includegraphics[width=\linewidth]{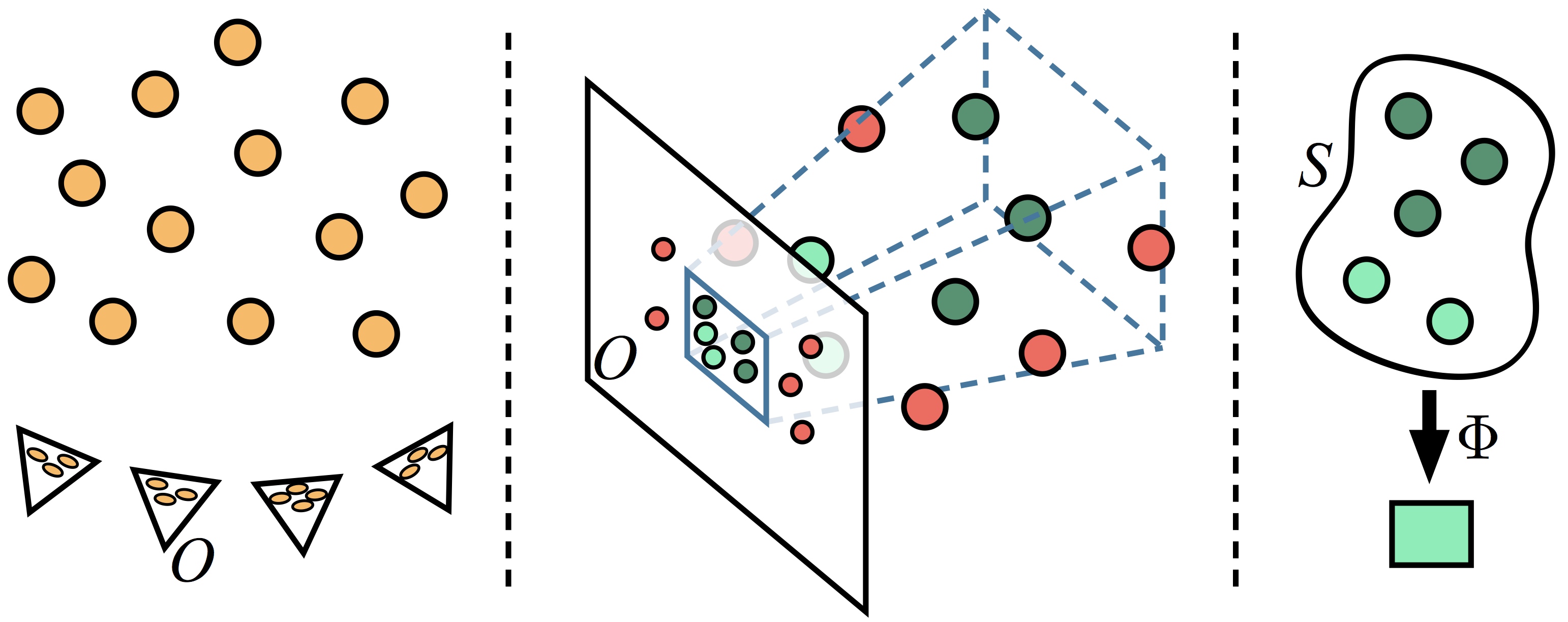}
	\caption{\label{fig:overview}Conceptual overview of our method. Samples are projected from video frames into scene-space (left), then a gathering step finds all samples that fall into the viewing frustum of an output pixel (middle), finally a filtering operation reduces this set to the output pixel's color (right).}
\end{figure}

The intuition behind our method can be summarized as follows, Fig.~\ref{fig:overview}.
For each output pixel, we gather all samples that lie within a 3D region defined by the pixel frustum of the output pixel, corresponding to all potential observations of the same scene point.
We then filter this sample set to generate an output pixel color by weighting the samples appropriately. 
This intuition is conceptual only; in practice the gathering step is performed directly in the images, avoiding the need for a costly intermediate 3D point cloud representation.

In the following, we write the color of a pixel $p$ at frame $f$ in the input video $\bm{I}$ as $\bm{I}^f(p)$.
Our goal is to compute all output pixel colors $\bm{O}^f(p)$.
For each $\bm{O}^f(p)$, we draw a set of samples $\bm{S}^f(p)$ directly from $\bm{I}$.
A sample $s \in \mathbb{R}^7$ is composed of color ($s_{rgb} \in \mathbb{R}^3$), scene-space position ($s_{xyz} \in \mathbb{R}^3$), and frame time ($s_{f} \in \mathbb{R}$).
\final{Color $s_{rgb}$ is in the range [0-255], $s_{xyz}$ are in scene space units, where the scene is scaled such that 90\% of the points lie in a $10^3$ cube, and $s_{f}$ is in units corresponding to the frames of the input video.}
Samples are generated by projecting a pixel from an input frame $\bm{I}^f$ using camera matrix $\bm{C}^f$ and its respective depth value $\bm{D}^f(p)$, Sec.\ref{sec:sampling}.
Filtering is defined as a function $\Phi(\bm{S}) \in \bm{\mathcal{P}}(\mathbb{R}^7) \rightarrow \mathbb{R}^3$ that takes a sample set and produces an output color per pixel, Sec.~\ref{sec:processing}.
In the following we drop the index $f$ for clarity when considering individual, sequentially processed frames.

\begin{figure}[t]
\setlength{\tabcolsep}{1px}
\centering
\def\myheighta{3.1cm}
\begin{tabular}{*{2}{c@{\hspace{3px}}}}
\includegraphics[height=\myheighta,clip,trim = 100 0 200 0]{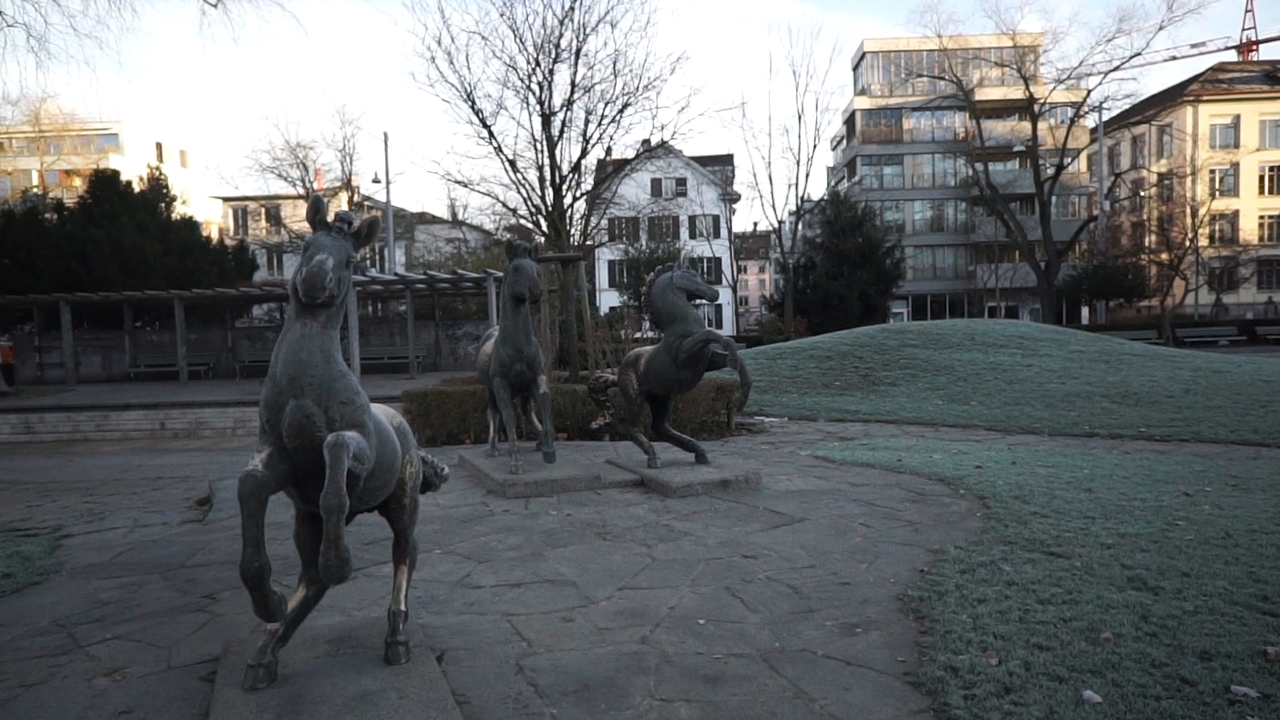} &
\includegraphics[height=\myheighta,clip,trim = 100 0 200 0]{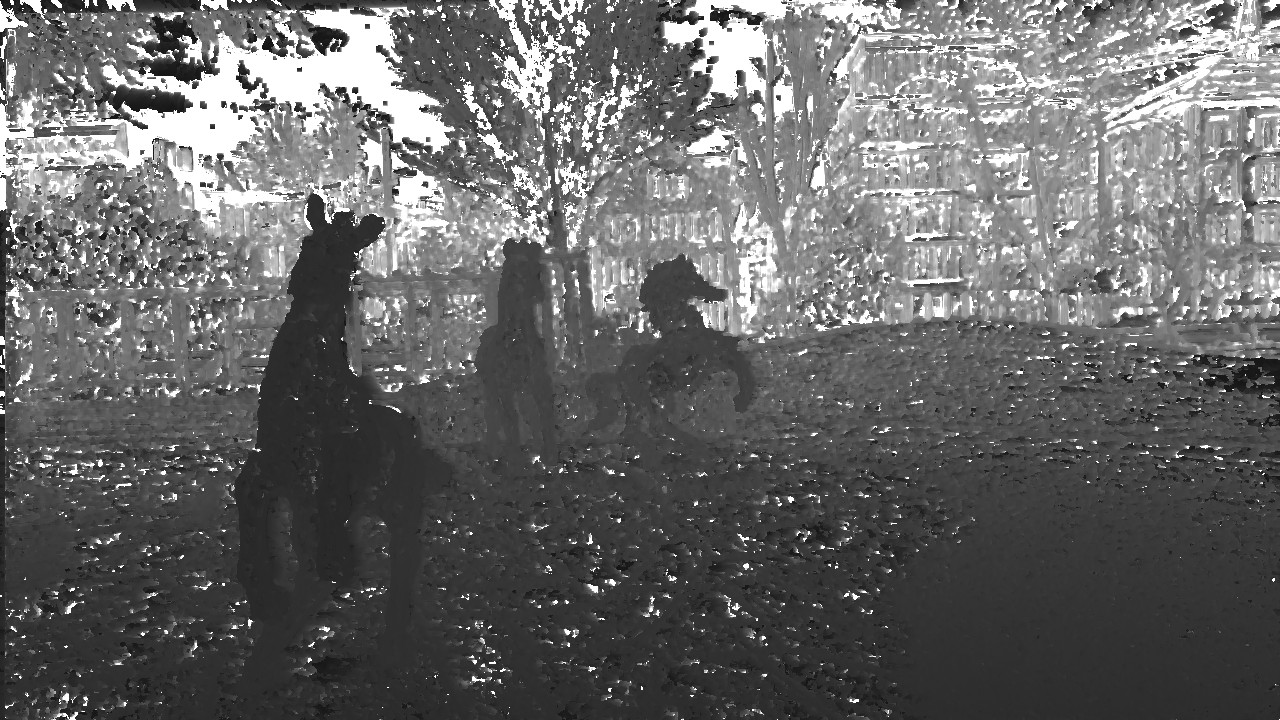} \\
Input & Depth map \\
\multicolumn{2}{c}{
\includegraphics[width=\linewidth,clip,trim = 0 100 0 100]{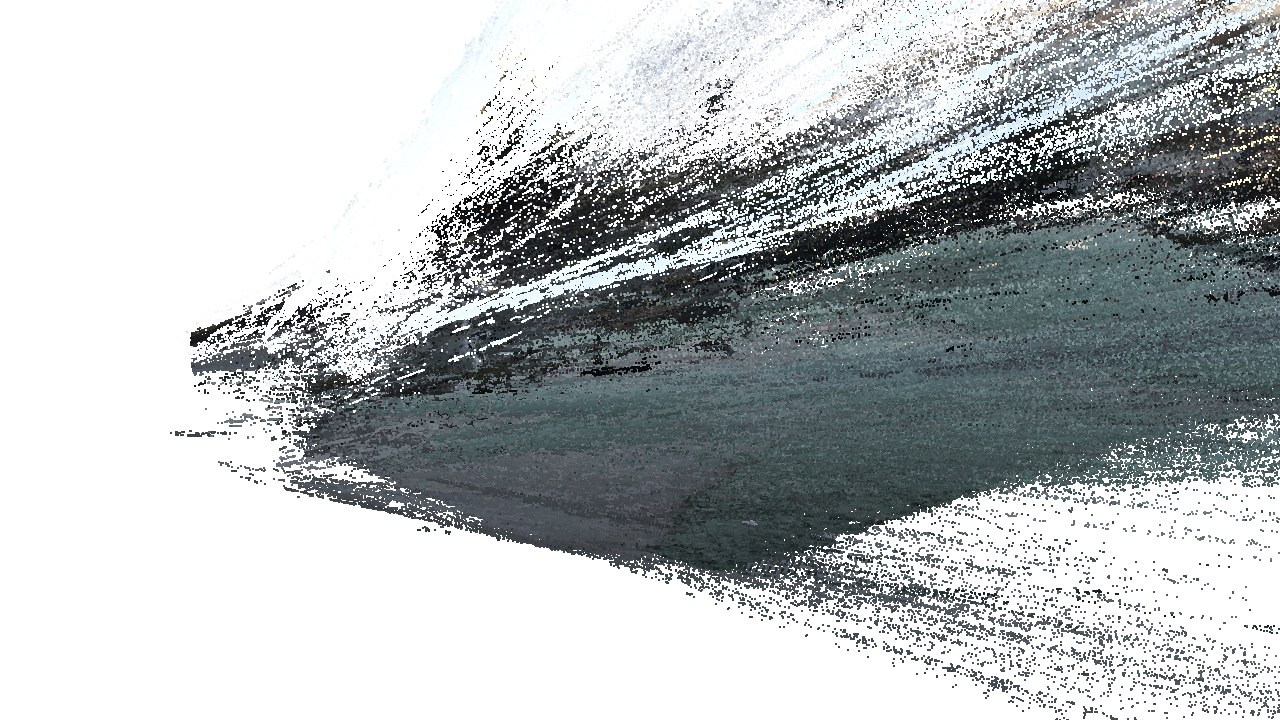}
}\\
\multicolumn{2}{c}{
Point cloud
}

\end{tabular}
\caption{\label{fig:input}Visualization of the quality of depth information our method is able to use. The point cloud below shows a side view of 5 images projected into scene-space. The amount of 3D outliers is clearly visible.}
\end{figure}

\paragraph{Preprocessing}
As a preprocessing step, we derive camera calibration parameters (extrinsics and intrinsics) $\bm{C}$, and depth information $\bm{D}$ from the input video $\bm{I}$.
Images are processed in an approximately linear color space by gamma correction.
Unless otherwise specified, we compute camera calibration parameters automatically using commonly available commercial tools (NUKEX 9.0v5).
Dense depth is either derived from the input video and camera calibrations using multi-view stereo techniques or, \final{in the case of the action shots example,} acquired by a Kinect sensor. %
Unless otherwise mentioned, we use a simple, local depth estimation algorithm where the standard multi-view stereo data-term~\cite{DBLP:conf/cvpr/SeitzCDSS06} is computed over a temporal window around each frame.
For each pixel this entails searching along a set of epipolar lines defined by $\bm{C}$, and picking the depth value with the lowest average cost (we use sum of squared RGB color differences on $3\times3$ patches).
This simple approach does not include any smoothness term and therefore does not require any complex global optimization scheme, making it easy to implement and efficient to compute.
As expected it yields many local depth outliers, introducing high-frequency ``salt-and-pepper'' noise in the depth map.
Fig.~\ref{fig:input} shows an example of a noisy point cloud corresponding to a depth map computed using this approach. 
Sec.~\ref{sec:implementation} discloses timing details of the depth computation step and all subsequent components of our method. 

%% file: sampling.tex
\begin{figure}[t]
	\includegraphics[width=\linewidth]{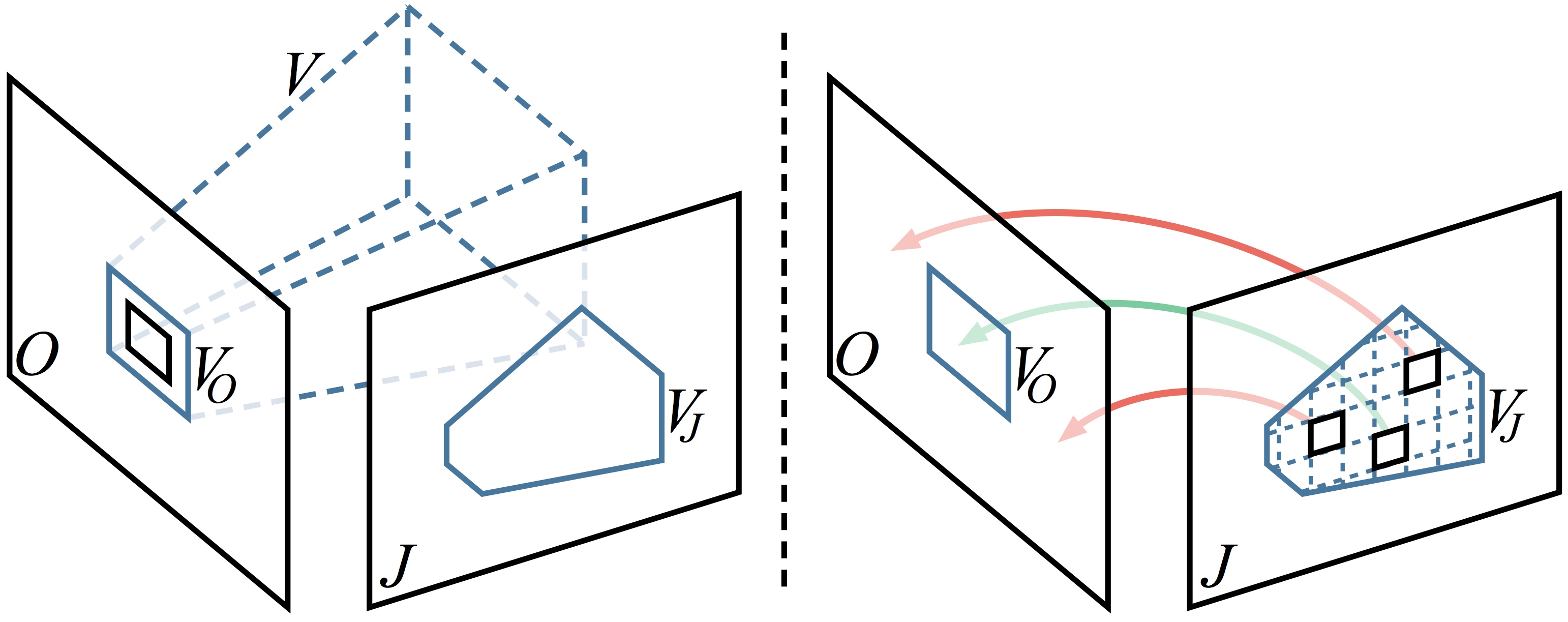}
	\caption{\label{fig:projection} 
	Sample gathering. 
	A pixel in output image $\bm{O}$ defines a frustum $\bm{V}$ which projects to a convex polygon $\bm{V_J}$ in an input frame $\bm{J}$ (left).
	Each pixel in $\bm{V_J}$ is then checked whether its projection in $\bm{O}$ is inside $\bm{V_O}$ (right).	
	Pixels that pass this test are added to the sample set. 
	The green arrow indicates a gathered sample, while the red arrows indicate samples that were tested, but rejected.
	}
\end{figure}

Next we discuss the general, application independent sample gathering step, describing how we compute a sample set $\bm{S}(p)$ for each output pixel $p$. 
The goal in constructing such a set is to collect multiple observations of the same \emph{scene point} visible to $p$.
For each output pixel $p$, a physical camera integrates information over a frustum-shaped 3D volume $\bm{V}$ in scene-space, Fig~\ref{fig:projection}.
Therefore the input video pixels that project into $\bm{V}$ are the \emph{samples} we want to collect.

The straightforward approach would be to project all video pixels into scene-space using their associated depth and camera matrices, and store the resulting 3D point cloud in a space partitioning structure (e.g., a kd-tree).
Gathering $\bm{S}(p)$ could then be done by querying the data structure.
However, a 1000-frame video at 720p resolution consists of nearly a billion 7D samples, and rendering an output video would require an equal number of \emph{frustum shaped} queries.
Computing this lookup on a general-purpose, data-agnostic data structure would be computationally intractable, so instead we exploit the underlying geometric nature of our input data to drastically boost efficiency.

The key idea behind our efficient gathering step is to exploit the duality between the scene and its 2D projections in the input video.
In order to find which pixels project into the frustum $\bm{V}$, we look at its projection into a single input frame $\bm{J}$.
All pixels that project into $\bm{V}$ must reside inside the respective 2D convex hull $\bm{V_J}$ (determined by projecting the frustum $\bm{V}$ into $\bm{J}$), Fig~\ref{fig:projection}.
Therefore, rather than storing and validating 3D points, we instead directly operate on the pixels in $\bm{V_J}$, looping through all $\bm{J}$.

More formally, given output camera matrix $\bm{C_O}$, the 3D frustum volume $\bm{V}$ of a pixel $p$ is simply defined as a standard truncated pyramid using the pixel location $(p_x, p_y)$ and a frustum size $l$:
\begin{equation}
\bm{V} = \{ \bm{C_O}^{-1} \cdot [p_x \pm \frac{l}{2}, p_y \pm \frac{l}{2}, \{near, far\}, 1]^T)\},
\end{equation}
where $near,far$ are the depth values of the near and far clipping planes (.01 and 1000 respectively).
The 2D frustum hull $\bm{V_J}$ is obtained by individually projecting the 3D frustum vertices of $\bm{V}$ into $\bm{J}$.
Because pixels inside of $\bm{V_J}$ may \emph{not} project into $\bm{V}$, but lie in front or behind it, we cannot simply accept the entire region.
Therefore, we rasterize all pixels $q \in \bm{V_J}$  and check whether their projection back into the output view lies within $\bm{V_O}$, Fig.~\ref{fig:projection}.
Specifically, we check the distance from $q$ projected back into $\bm{O}$ to the original pixel $p$.
\begin{equation}
\| p - C_O \cdot C_J^{-1} \cdot [q_x, q_y, q_d, 1]^T \|_1 < \frac{l}{2}
\label{eqn:sampletoworld}
\end{equation}
Each pixel $q$ passing this test is converted into a 7D sample and added to the sample set $\bm{S}(p)$.

In case of error-free depth maps, camera poses, and a static scene, the samples inside that pixel's frustum ($l=1$) would be a complete set of all observations of that scene point (as well as any occluded scene points). 
However, inaccuracies in camera pose and depth inevitably lead to false positives, i.e., outlier samples wrongly gathered, and false negatives, i.e., scene point observations that are missed.
To account for depth and camera calibration inaccuracies, we increase the per-pixel frustum size $l$ to cover a wider range. 
In most cases, we use $l=3$ pixels.
In general, this produces on the order of a few hundred to a thousand samples per set.
Note that the depth of the current output pixel $p$ is irrelevant at this stage of the algorithm.

Since output sample sets are handled independently, we can compute all frustum volumes for one output frame concurrently (i.e., the approach is parallelized over output pixels $p$).
This kind of local, independent parallelism is especially well suited for GPU implementation.
To handle the large volumes of input data, and to additionally maximize data coherency during the computation, each pixel in $\bm{O}$ gathers all samples from $\bm{I}^f$ in parallel before moving on to $\bm{I}^{f+1}$.

%% file: processing.tex
Now that we have computed the sample set $\bm{S}(p)$, the second step is to determine a final pixel color $\bm{O}(p)$ from this sample set.
Among the samples in $\bm{S}(p)$, some will correspond to a scene point observed by the output camera, but others will come from occluded regions, incorrect 3D information, or moving objects.
An application specific weighting function is used to emphasize samples that we can trust, while reducing the influence of outliers.
All results we show in the following are computed using a sample set filtering operation of the form:
\begin{equation}
\bm{O}(p) = \Phi(\bm{S}(p)) = \frac{1}{\bm{W}}{\sum_{s \in \bm{S}(p)}{w(s) s_{rgb}}}
\label{eq:filtering}
\end{equation}
where $w(s)$ is an application specific weighting function and $\bm{W} = \sum_{s \in \bm{S}(p)}{w(s)}$, is the sum of all weights.
One key property of Eq.~\ref{eq:filtering} is that it is very fast to compute. 
This allows us to take advantage of the large amount of redundant data present in video.
To give an intuition for the importance of efficiency; for a 30 second 720p video at 30fps, and number of samples per set $|\bm{S}(p)| \approx 1000$, the output video will be created from about a \textbf{trillion} weighted samples.
We show how this can be done in a reasonable amount of time on a desktop computer.
In empirical tests, we observe that, on average $\frac{\bm{W}}{|\bm{S}(p))|} \approx .18$, \final{indicating that around 82\% of the information in $\bm{S}(p)$ is discarded by our approach.}

A central feature of our method is the flexibility it provides in defining different weighting functions $w(s)$ on the 7D samples. 
In particular, it is  straightforward to specify effects based on \emph{scene-space} coordinates by making $w(s)$ depend on the scene-space position of a sample.
We demonstrate this in several applications, such as action shots, and inpainting defining approximate 3D regions (bounding boxes), each with their own set of parameters for $w(s)$.

We demonstrate the general applicability of our method by first showing a number of difficult fundamental video processing operations, followed by advanced video effects.
All results were computed using different variants of the weighting function $w(s)$.
We encourage the reader to watch the accompanying video to assess the results.
\final{All video results and datasets are available on the project website.}

%% file: applications.tex
\begin{figure*}[t]

\setlength{\tabcolsep}{1px}
\centering
\newcommand{\denoiseinclude}[1]{\includegraphics[height=3.4cm,trim = 200 100 200 100,clip]{#1}}
\newcommand\beginmytikz[1]{\begin{tikzpicture}[spy using outlines={magnification={#1},width=2.75cm, height=2.8cm,connect spies}]}
\newcommand\myspy[4]{\spy [red] on ({#1},{#2}) in node [below] at ({#3},{#4})}

\begin{tabular}{*{4}{c@{\hspace{0px}}}}
{\begin{sideways}\parbox{5cm}{\centering Denoising}\end{sideways}} &
\beginmytikz{4}
	\node [] { \denoiseinclude{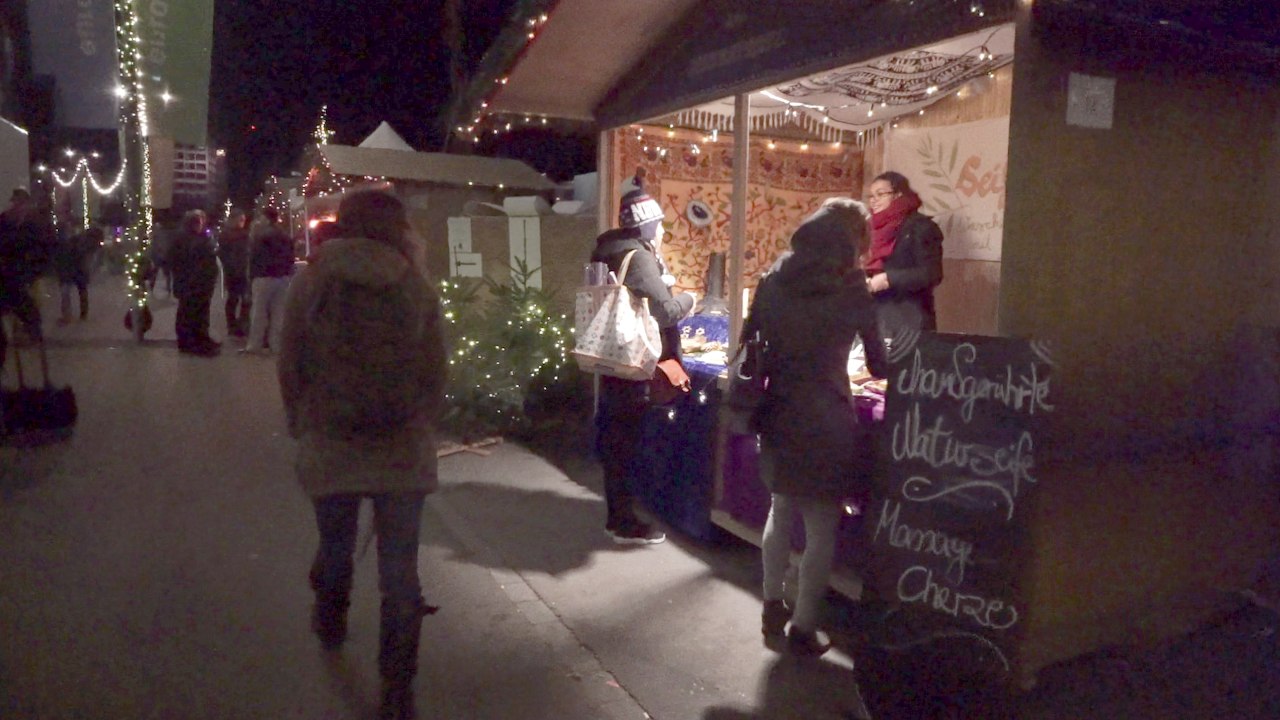} };
	\myspy{-1.2}{.15}{-1.5}{-1.8};
	\myspy{2.15}{1}{1.5}{-1.8};	
\end{tikzpicture} &
\beginmytikz{4}
	\node [] { \denoiseinclude{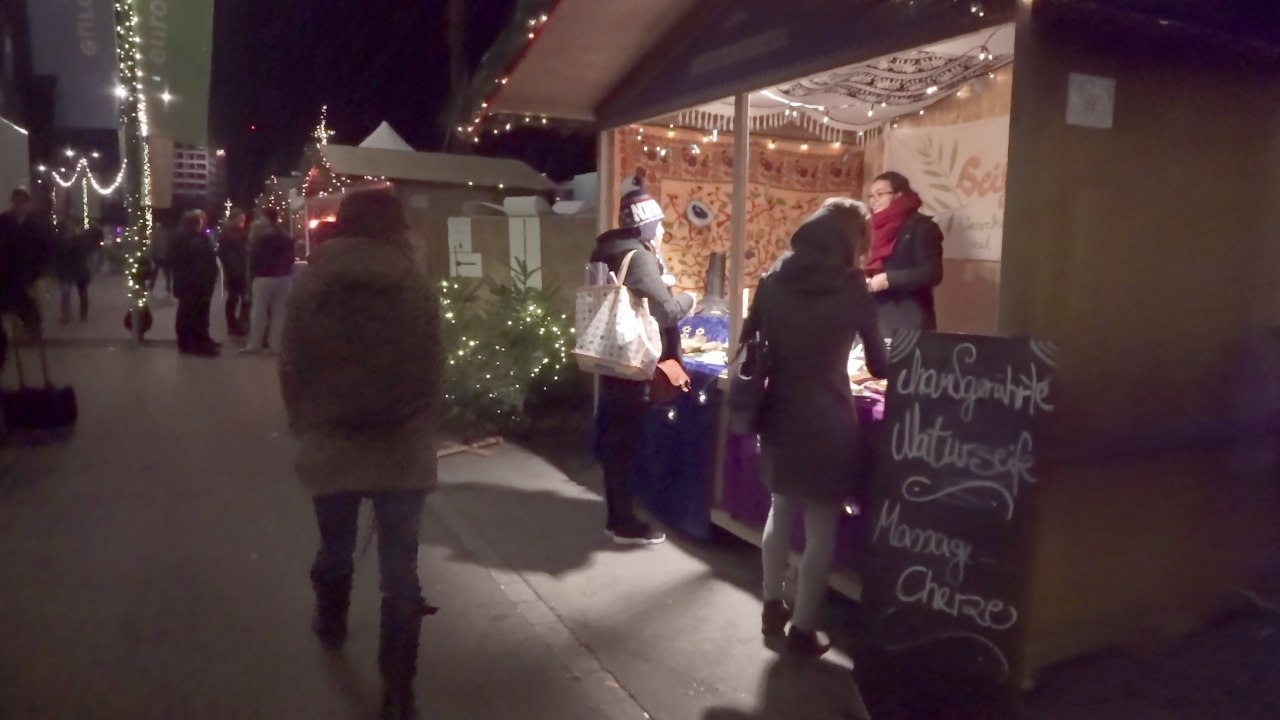} };
	\myspy{-1.2}{.15}{-1.5}{-1.8};
	\myspy{2.15}{1}{1.5}{-1.8};	
\end{tikzpicture} & 
\beginmytikz{4}
	\node [] { \denoiseinclude{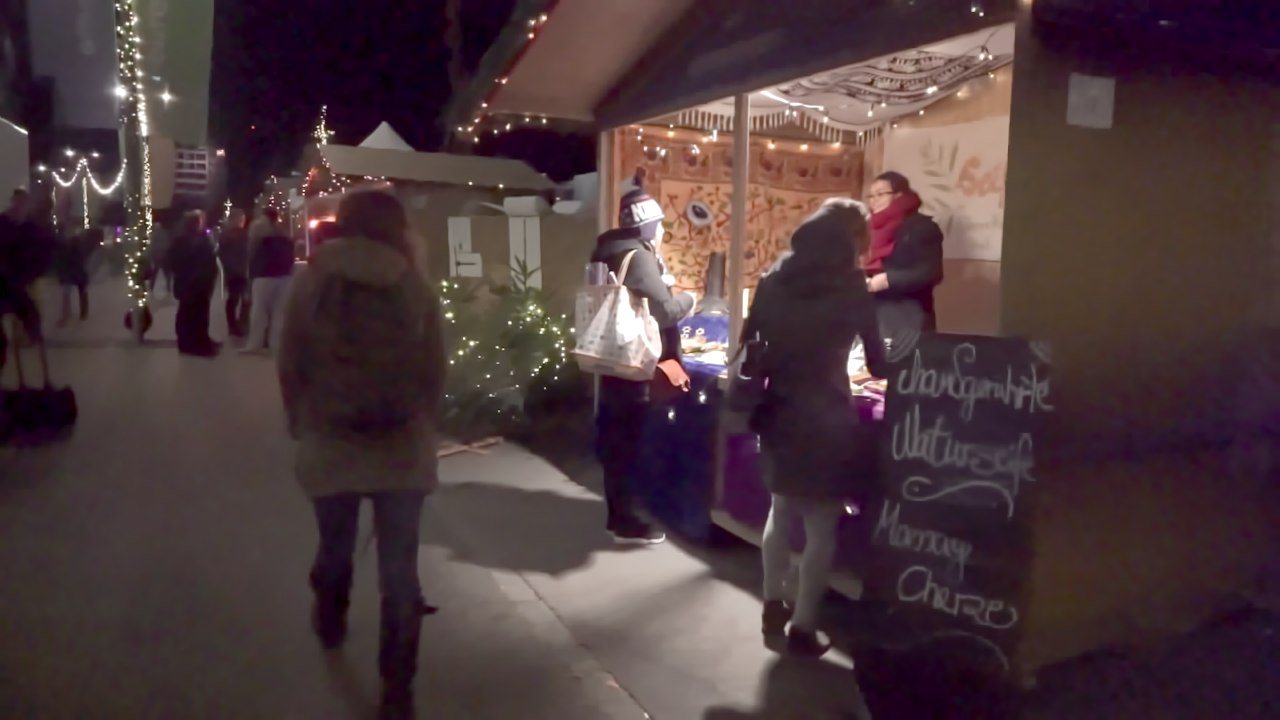} };
	\myspy{-1.2}{.15}{-1.5}{-1.8};
	\myspy{2.15}{1}{1.5}{-1.8};	
\end{tikzpicture} \\    
& Input & Scene-space denoising & BM3D~\protect\cite{DBLP:journals/tip/DabovFKE07} \\
\end{tabular}

\def\myheight{4cm}
\begin{tabular}{*{4}{c@{\hspace{5px}}}}
{\begin{sideways}\parbox{\myheight}{\centering Deblurring}\end{sideways}} &
\includegraphics[clip,trim=450 400 450 50,height=\myheight]{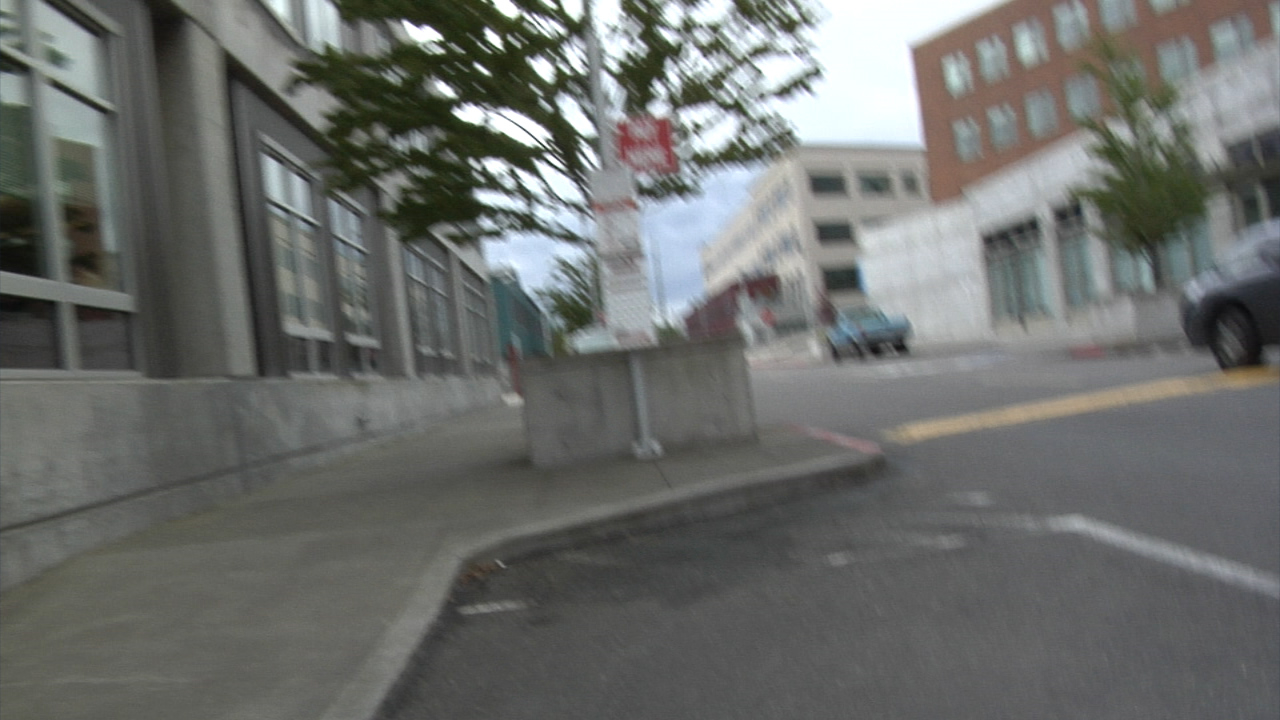} &
\includegraphics[clip,trim=450 400 450 50,height=\myheight]{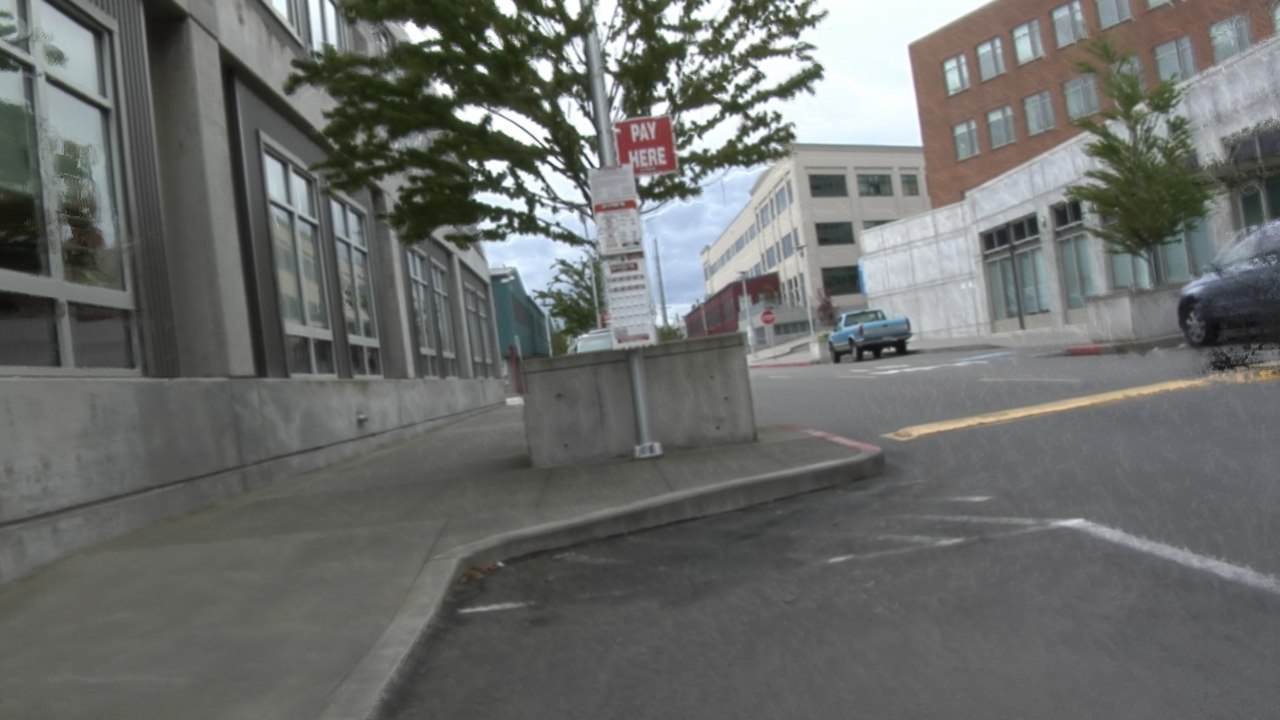} &
\includegraphics[clip,trim=450 400 450 50,height=\myheight]{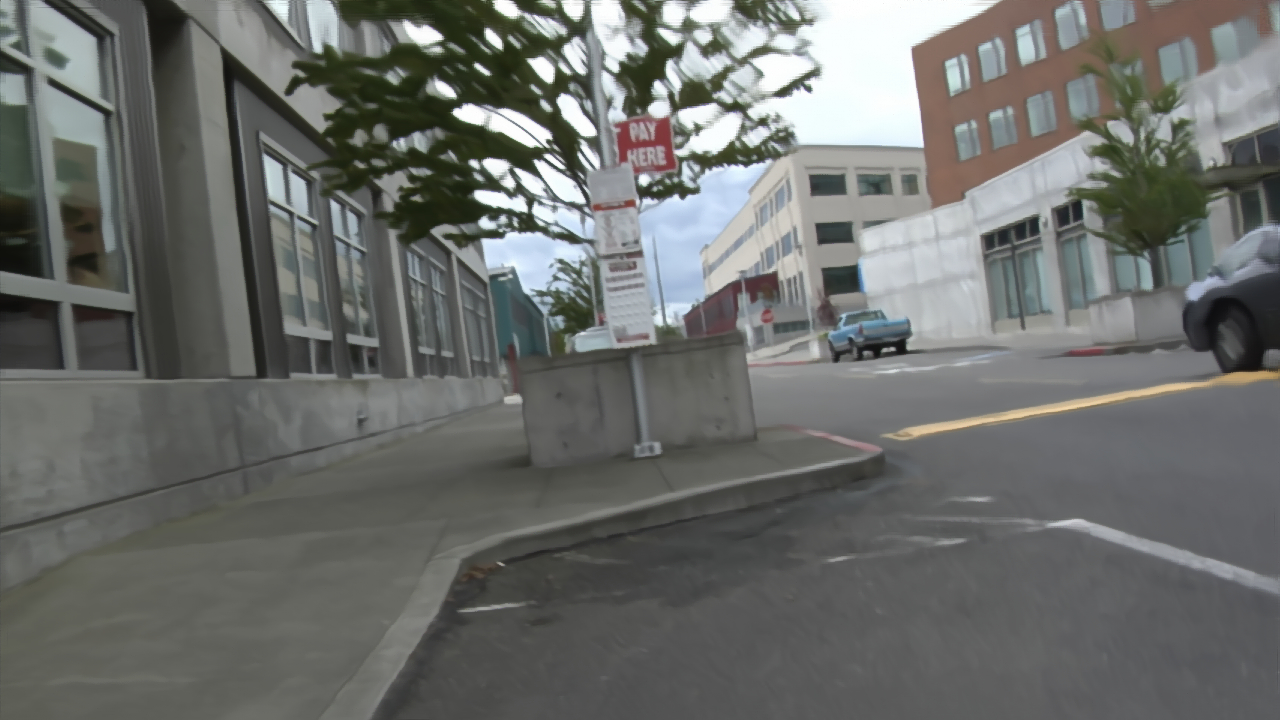} \\
& Input & Scene-space deblurring & \protect\cite{DBLP:journals/tog/ChoWL12}  \\
\end{tabular}

\caption{\label{fig:fundamental}Results of our method on video denoising and deblurring. We obtain similar quality results to existing state-of-the-art approaches specifically tailored to the respective task.}
\end{figure*}

\begin{figure*}[t]
\setlength{\tabcolsep}{1px}
\centering
\def\myheight{3.9cm}

\newcommand{\superinclude}[1]{\includegraphics[height=\myheight,trim = 2000 1000 1400 850,clip]{#1}}
\begin{tabular}{*{4}{c@{\hspace{5px}}}}
{\begin{sideways}\parbox{\myheight}{\centering Super resolution}\end{sideways}} &
\superinclude{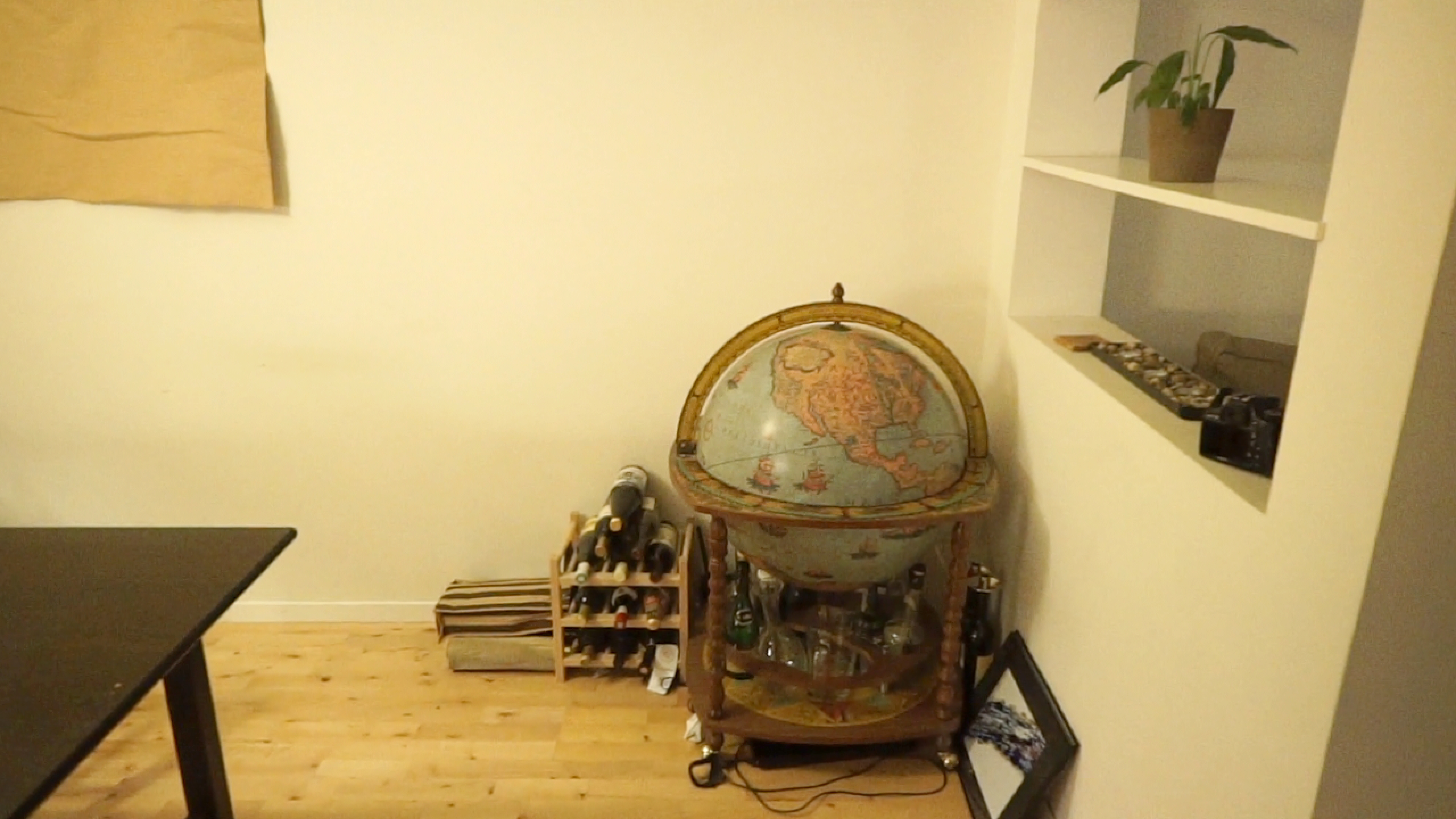} &
\superinclude{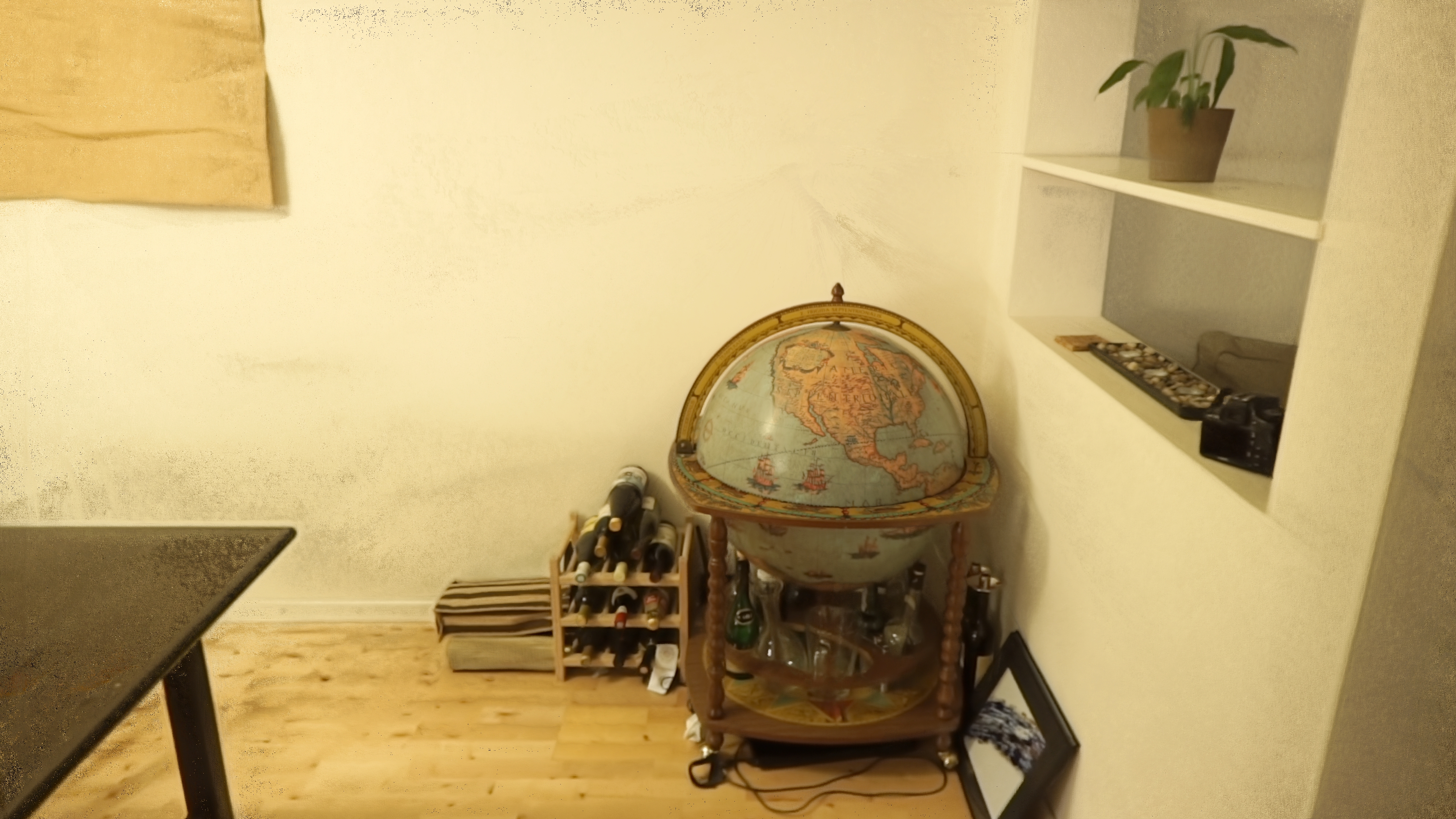} & 
\superinclude{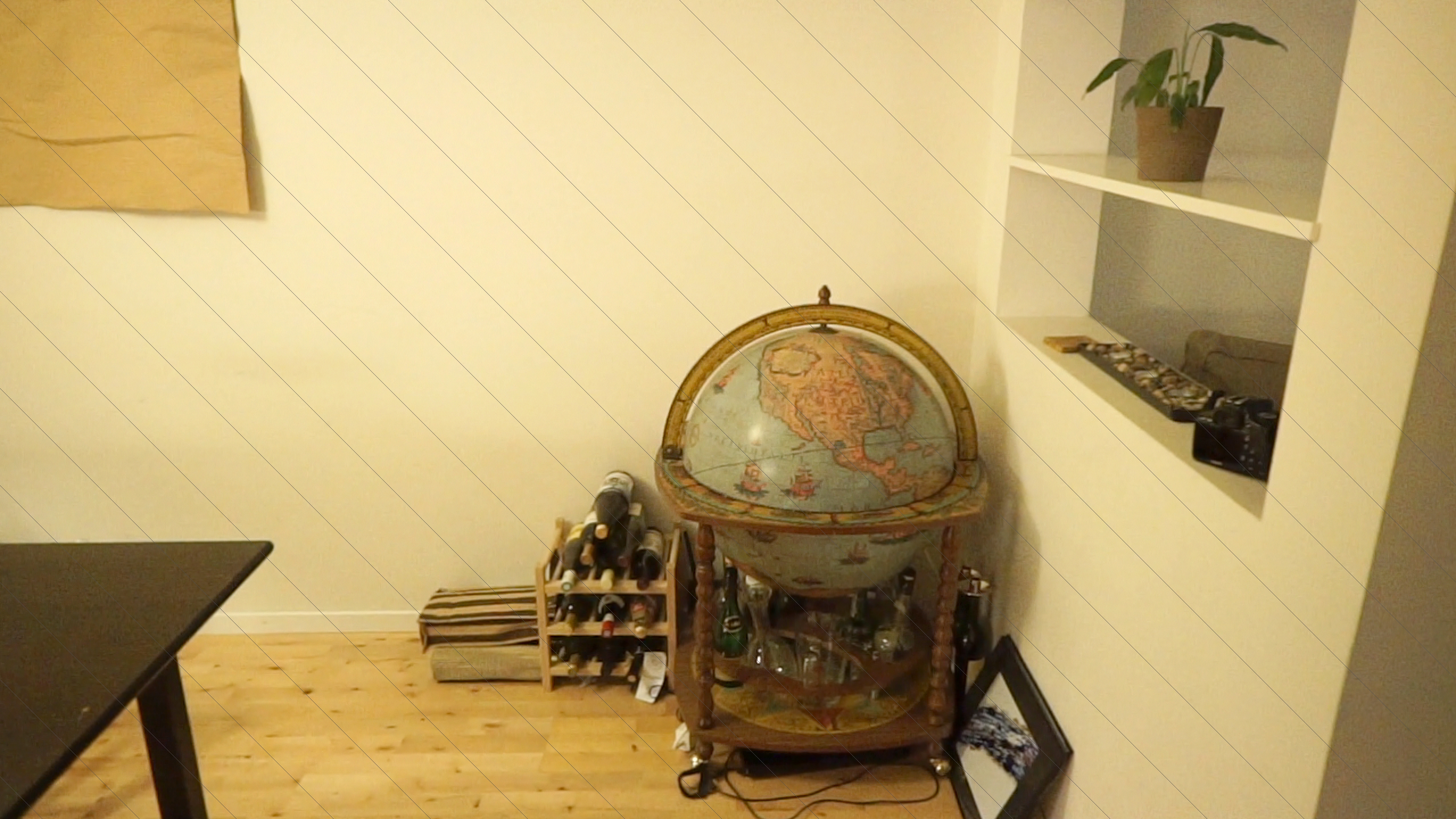} \\
& Input & Scene-space super resolution & Infognition super resolution~\protect\shortcite{Infognition} \\
\end{tabular}

\caption{\label{fig:fundamental2}A comparison using our framework and a commercial tool for super-resolution.}
\end{figure*}

\paragraph{Denoising}
\label{sec:application:denoise}
As the same scene point is observed many times throughout a video, we can use these multiple observations to denoise input frames.
Simply averaging all samples in $\bm{S}(p)$ by setting $w(s) = 1$ causes occluded and noisy samples to corrupt the result, Fig.~\ref{fig:mean}.
One key observation is that we have a reasonable prior on the expected color $\bm{O}^f(p)$ of an output pixel $p$, namely the input pixel color $\bm{I}^f(p)$ at the same frame and pixel location.
We call the sample that originates from projecting $\bm{I}^f(p)$ into scene-space to be the \emph{reference} sample $s_{ref}$.
Filtering can then be done as a weighted sum of samples, where weights are computed as a multivariate normal distribution with mean $s_{ref}$:
\begin{equation}
w_{denoise}(s) = \exp \left( {-\frac{(s_{ref}-s)^2}{2\sigma^2}} \right) .
\label{eq:denoising}
\end{equation}
While we use the above notation for clarity, samples are actually represented in a 7D space and we use a diagonal covariance matrix.
We call the diagonal entries $\sigma_{rgb}$ for the three color dimensions, $\sigma_{xyz}$ for the scene-space position and $\sigma_{f}$ for the frame time.
For denoising, we use typical parameters $\sigma_{rgb} = 40$,  $\sigma_{xyz} = 10$, $\sigma_{f} = 6$.
\final{
The values of $\sigma$ are set based on the expected variance of their respective modalities; due to the low quality of the input videos, depth estimates in this application are very noisy and $\sigma_{xyz}$ is set to a high value (scenes are scaled to a $10^3$ cube).
}
For some applications we do not list all three parameters, denoting that some dimensions are not used. %

We compare the result of our scene-space filter with a state-of-the-art video denoising method (BM3D)~\cite{DBLP:journals/tip/DabovFKE07} in Fig.~\ref{fig:fundamental}. 
Despite a simple filtering operation, we achieve similar quality denoising results, demonstrating the power of working in scene-space with large volumes of data.
Our method even produces reasonable results for scenes exhibiting lighting changes and limited amounts of foreground motion.
We show examples of these in the supplemental video.

\begin{figure}[t]
\begin{tikzpicture}
	\begin{axis}[
		height=4.5cm,
		width=\linewidth,
		xlabel=Standard deviation of noise added,
		ylabel=PSNR]
	\addplot[color=red,mark=o,mark options={scale=0.6}] coordinates {
		(2,41.85)
		(4,36.12)
		(8,30.20)
		(16,24.27)
		(32,18.48)
		(64,13.17)
	};
	\addlegendentry{Noisy input}	
	\addplot[color=green,mark=o,mark options={scale=0.6}] coordinates {
		(2,42.23)
		(4,37.81)
		(8,33.05)
		(16,28.07)
		(32,22.96)
		(64,18.00)
	};
	\addlegendentry{BM3D}	
	\addplot[color=blue,mark=o,mark options={scale=0.6}] coordinates {
		(2,41.53)
		(4,38.30)
		(8,35.70)
		(16,31.23)
		(32,27.51)
		(64,22.70)		
	};	
	\addlegendentry{Scene-space}	
	\end{axis}
\end{tikzpicture}
\caption{\final{PSNR for denoising. Synthetic noise was removed from a real video sequence using BM3D and our scene-space method.}}
\label{fig:denoisingPSNR}
\vspace{-2mm}
\end{figure}
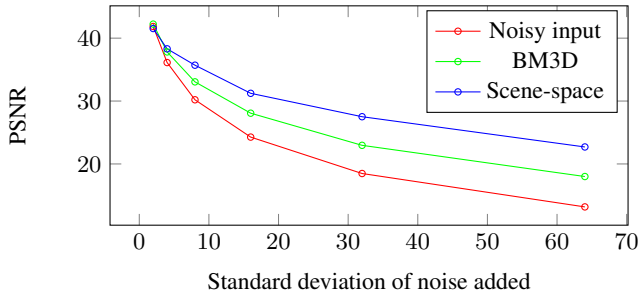

\final{
We conducted an additional quantitative evaluation to compare to spatiotemporal BM3D.
In this experiment, we took a largely noise-free video sequence, and added zero mean, grayscale Gaussian noise with varying standard deviations. 
After adding increasing amounts of noise, we ran the noisy videos through our entire automatic pipeline, including depth computation, camera calibration, and denoising.
Fig.~\ref{fig:denoisingPSNR} shows the resulting plot of PSNR values.
In the case of a static scene and camera motion, we observe that our method is able to outperform BM3D by incorporating 3D information in the filtering.
}

\paragraph{Deblurring}
We can also use our method for deblurring video frames that are blurry due to sudden camera movements, e.g., during hand-held capture.
We use the same equation as above, but add to it a measure of frame blurriness:
\begin{equation}
w_{deblur}(s) =  \exp \left( {-\frac{(s_{ref}-s)^2}{2\sigma^2}} \right) \sum_{q \in \bm{I}^{s_f}}{\left|\nabla \bm{I}^{s_f}(q) \right|} ,
\label{eq:deblurring}
\end{equation}
where $\nabla$ is the gradient operator, and ${I}^{s_f}$ is the frame that the sample $s$ originated from.
The first part is the same multivariate normal distribution as Eq.(\ref{eq:denoising}), and blurriness is computed as the sum of gradient magnitudes in ${I}^{s_f}$.
This down-weights the contribution from blurry frames when computing an output color. 
We use values $\sigma_{rgb} = 200$,  $\sigma_{xyz} = 10$, $\sigma_{f} = 20$. 
We compare to a recent state-of-the-art approach that uses a patch search in space and time to find non-blurred patches that combines the content of these patches with the current frame~\cite{DBLP:journals/tog/ChoWL12}.
We can observe similar quality results, Fig~\ref{fig:fundamental}.

\begin{figure*}[t]
\setlength{\tabcolsep}{1px}
\centering
\def\myheight{3.9cm}

\def\rmyheight{3.5cm}
\begin{tabular}{*{5}{c@{\hspace{3px}}}}
{\begin{sideways}\parbox{\rmyheight}{\centering Object semi-transparency}\end{sideways}} &
\includegraphics[height=\rmyheight,trim = 400 100 290 120,clip]{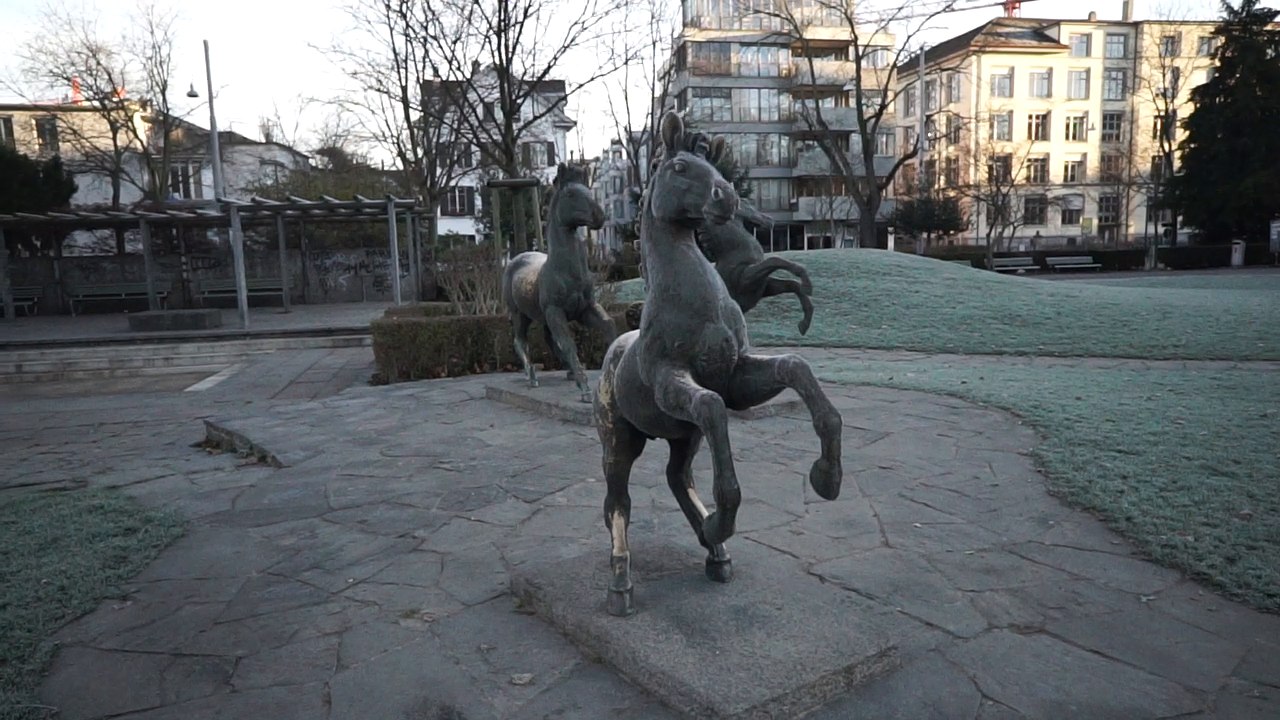} &
\includegraphics[height=\rmyheight,trim = 75 0 175 0,clip]{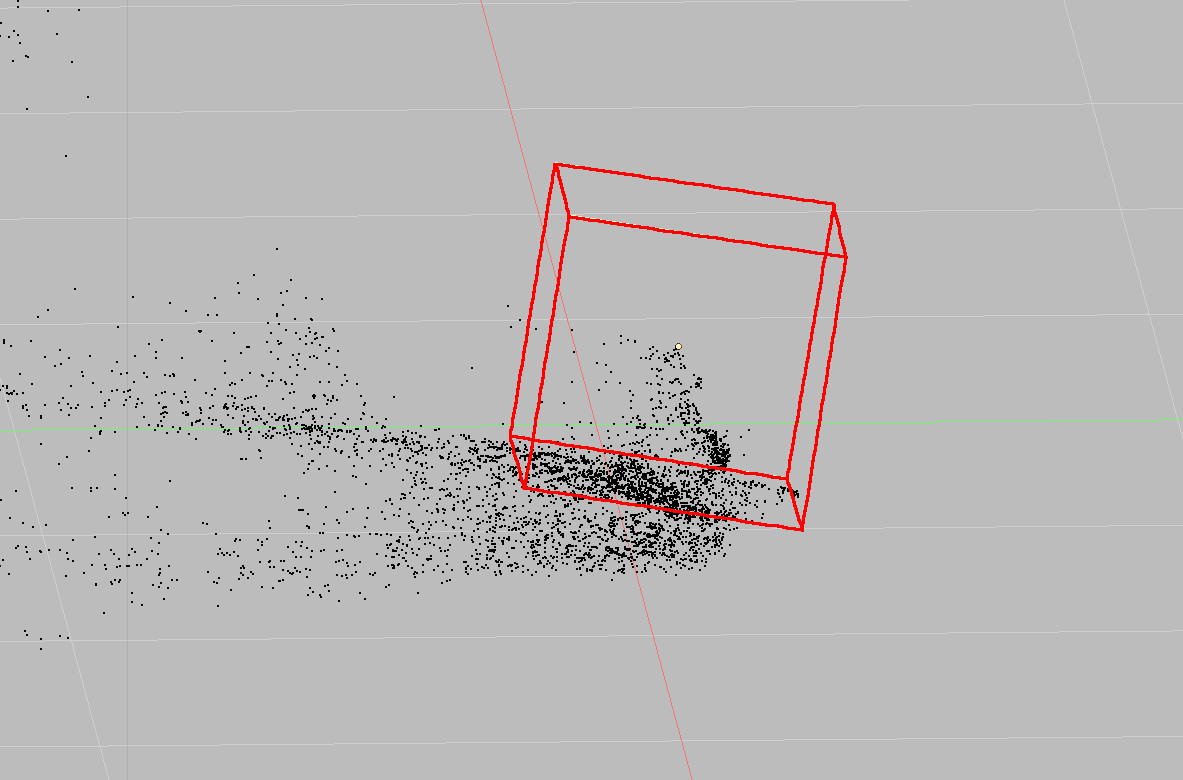} &
\includegraphics[height=\rmyheight,trim = 400 100 290 120,clip]{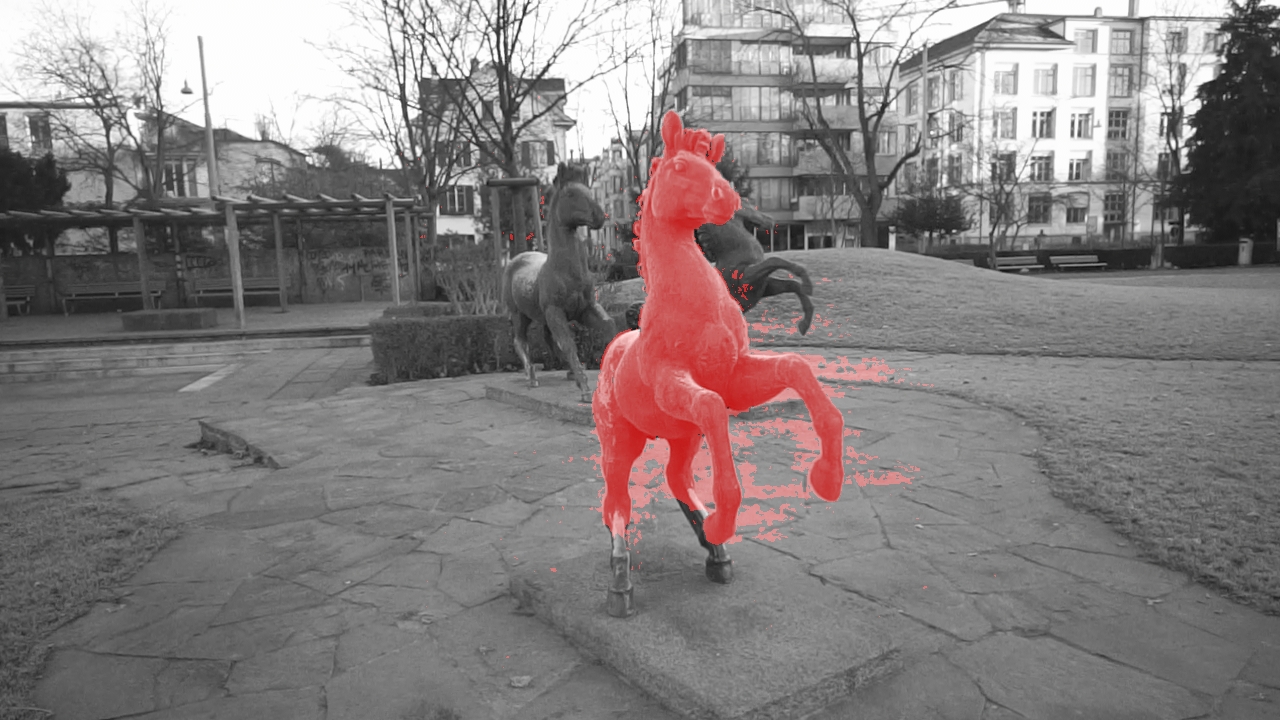} &
\includegraphics[height=\rmyheight,trim = 400 100 290 120,clip]{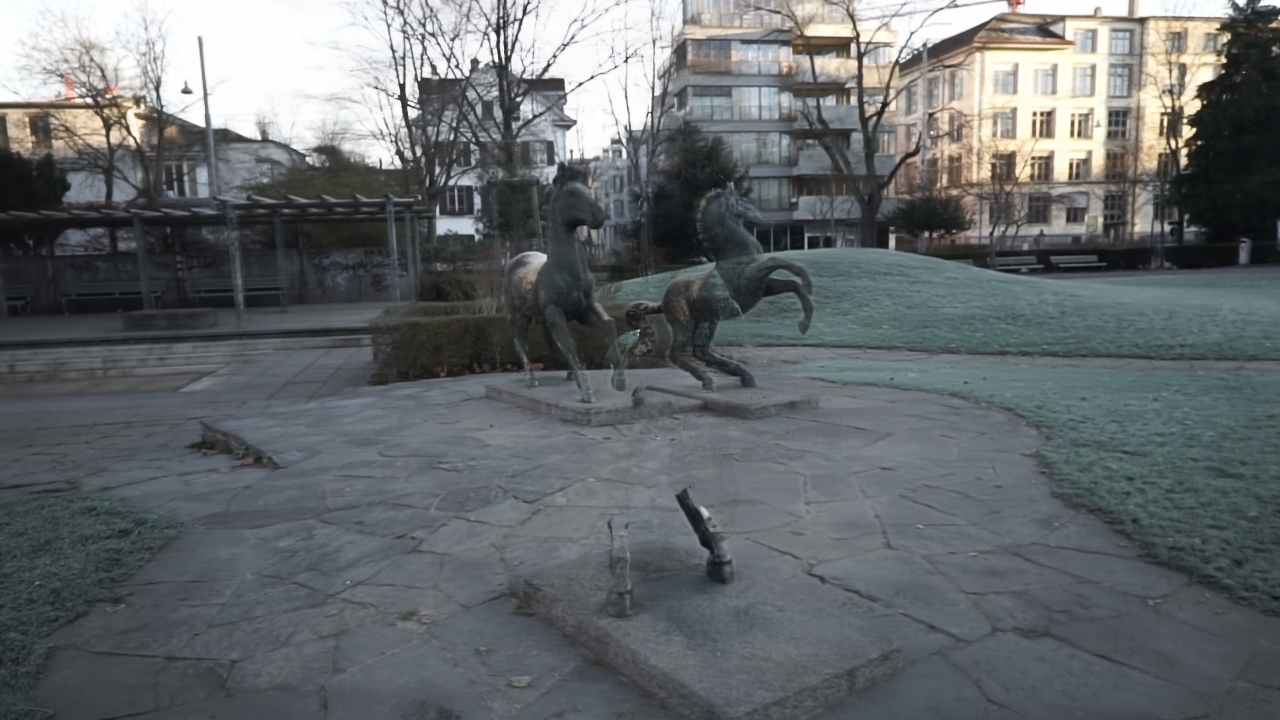} \\
& Input & 3D mask & 3D mask reprojected & Scene-space semi-transparency \\
\end{tabular}

\newcommand{\inpaintinclude}[1]{\includegraphics[height=3cm,trim = 200 80 240 20,clip]{#1}}
\newcommand{\closeinpaintinclude}[1]{\includegraphics[height=3cm,trim = 430 200 290 120,clip]{#1}}
\begin{tabular}{*{5}{c@{\hspace{3px}}}}
{\begin{sideways}\parbox{3cm}{\centering Video inpainting}\end{sideways}} &
\inpaintinclude{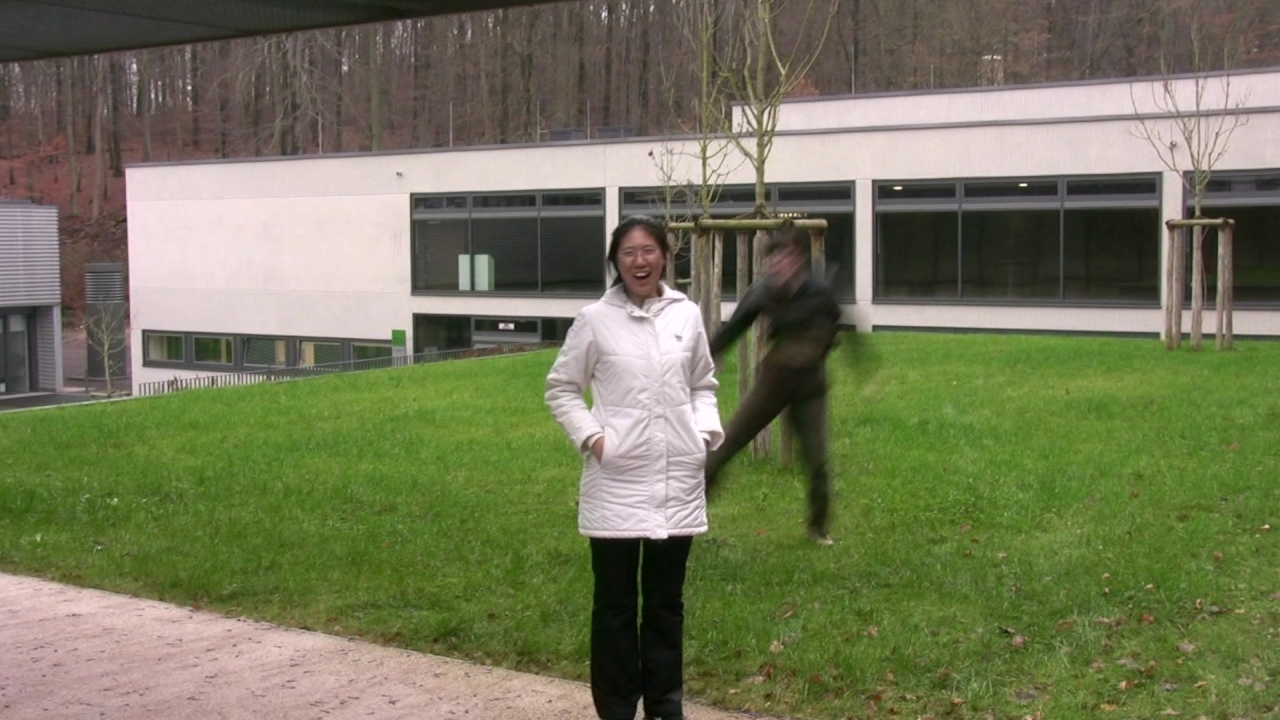} &
\inpaintinclude{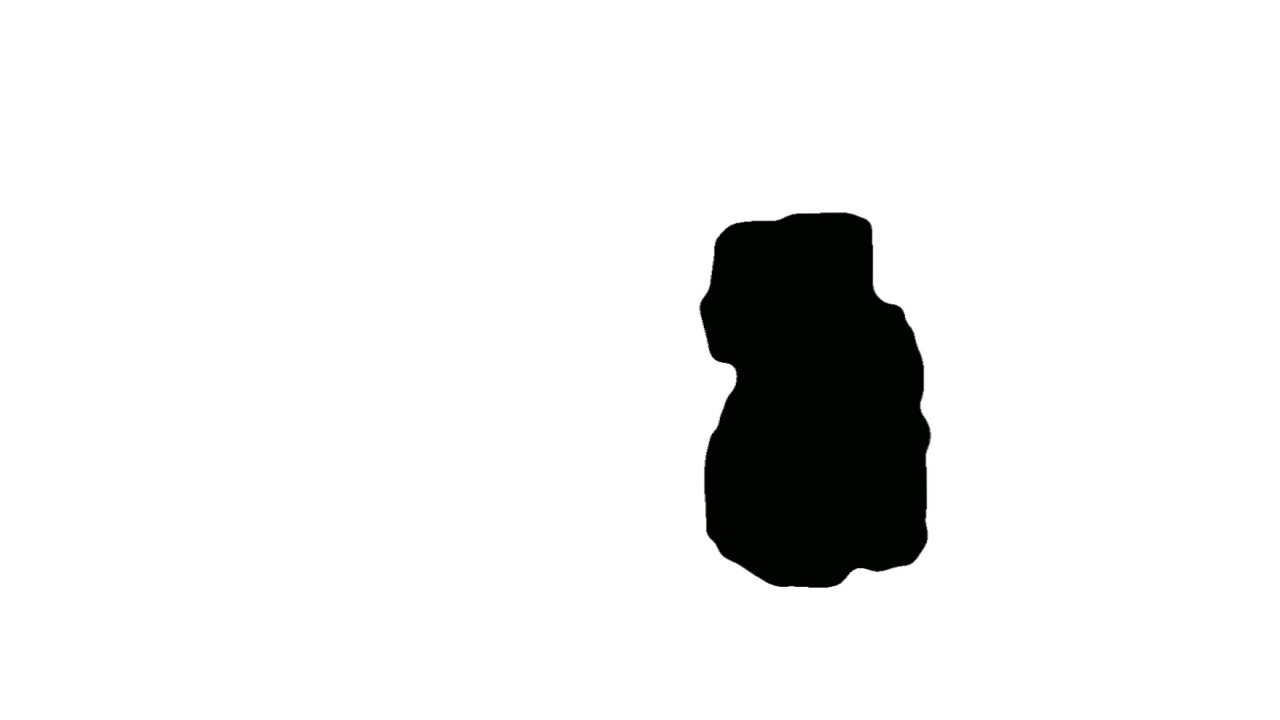} &
\closeinpaintinclude{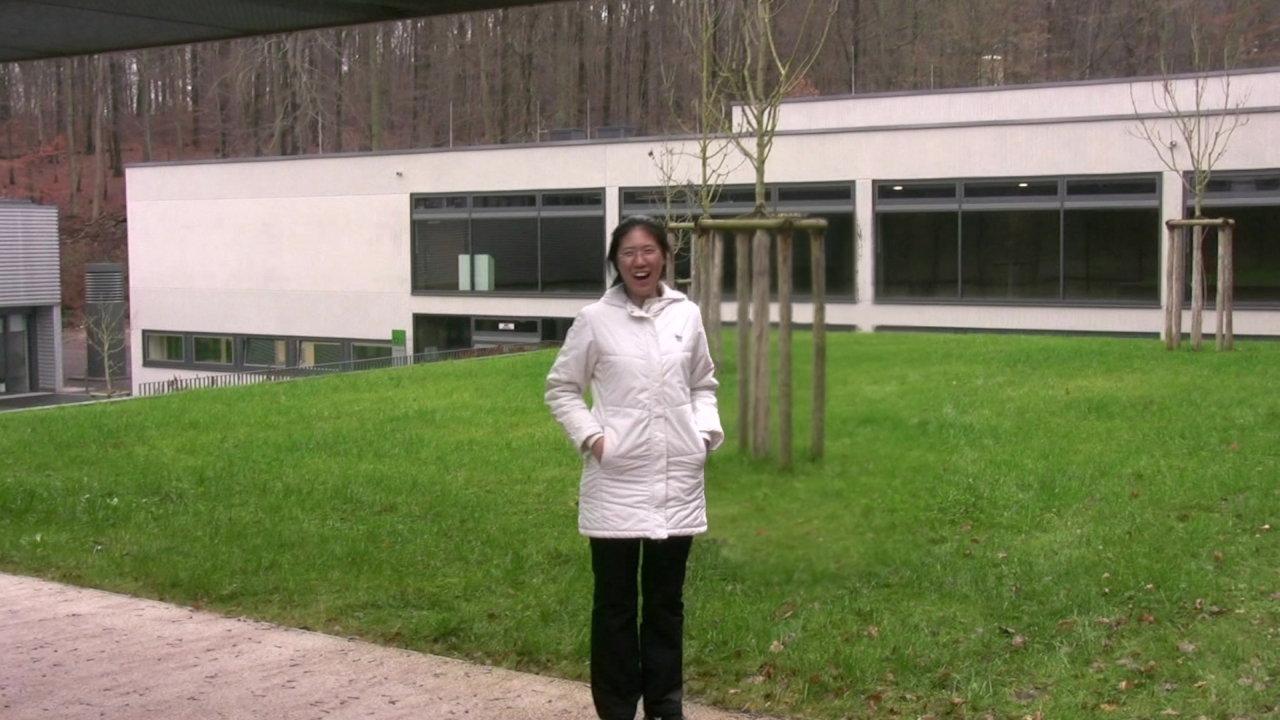} &
\closeinpaintinclude{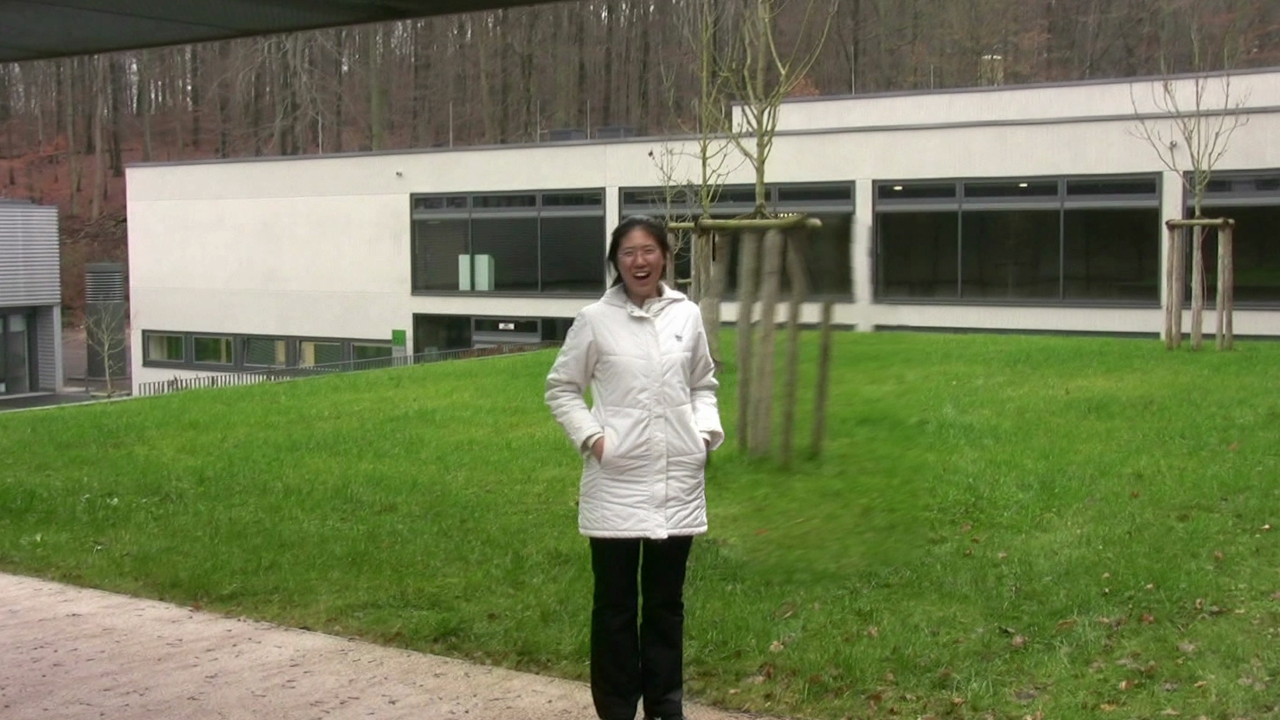} \\
& Input & Background mask & Scene-space inpainting & \protect\cite{DBLP:conf/eccv/GranadosKTKT12} \\
\end{tabular}	
	\caption{Scene-space samples can be used for video inpainting. Regions can be specified either in scene-space (top row) or in image-space (bottom row). The 2D background mask is inverted for clarity. %
	}
	\label{fig:inpainting}
\end{figure*}

\paragraph{Super resolution}
\label{sec:application:superresolution}
We can also perform a scene-space form of super resolution, where the goal is to create a high-resolution output video $\widehat{\bm{O}}$ from a low-resolution input video $\bm{I}$.
The traditional approach of using sub-pixel shifts derived from aligning multiple images, and solving for the high-resolution image, e.g., using iterative reweighted least squares~\cite{DBLP:journals/tvcg/SunkavalliJKCP12}, or using external priors, e.g., on image gradients~\cite{DBLP:conf/cvpr/SunXS08}.
Instead, we simply choose a weighting scheme that prefers observations of the scene point with the highest available resolution.
We assume that each scene point is most clearly recorded when it is observed from as close as possible (i.e., the sample with the \emph{smallest} projected area in scene space).
To measure this, we introduce a new sample property that we call $s_{area}$.
The scene-space area of a sample is computed by projecting its pixel corners into the scene and computing the area of the resulting quad. 
Assuming square pixels it is sufficient to compute the length of one edge. Let $p_l$ and $p_r$ be the left and right edge pixel locations of a sample located at $p$ and $\bm{C}$ be the camera matrix for the sample's frame $s_f$;
\begin{equation}
s_{area} = \big\| \bm{C}^{-1} \cdot [p_l, D(p),1]^T  - \bm{C}^{-1} \cdot [p_r, D(p),1]^T \big\|_2^2
\end{equation}
We then use the following weighting function:
\begin{equation}
w_{sr}(s) =  
\exp \left({-\frac{(s_{ref}-s)^2}{2\sigma^2}}\right)
\exp \left(-\frac{s_{area}}{2\sigma_{area}}^2\right) .
\label{eq:superresolution}
\end{equation}
We use parameters values $\sigma_{rgb}=50$, and $\sigma_{area}=.02$.
Intuitively, the latter term down-weights samples that were observed by cameras from farther away, preferring the samples with more detailed information.
In order to generate reference samples $s_{ref}$ in this case, we bi-linearly upsample $\bm{I}$ to the output resolution. 
As our gathering step allows samples to be gathered from arbitrary pixel frustums, we simply gather samples from frustums corresponding to pixel coordinates from $\widehat{\bm{O}}$, rather than $\bm{O}$. 

\ignore{
	\paragraph{HDR rendering}
	\label{application:hdr}
	Can we combine lighting change (or aperture change) over the video, and use this to get a HDR output video?!
	Assuming we know the exposure change function, we can compute the corrected hdr color value for all scene-space locations where we have enough samples. We might even get a scene-space localized color lookup table, that can map from the recorded color and exposure function value to the "correct" color.
	How do we calibrate this?
	HDR
	\begin{itemize}
	\item \cite{DBLP:journals/tog/KalantariSBDGS13}: Patch based HDR Alignment, for video, aligning neighboring frames
	\item \cite{DBLP:conf/pg/MertensKR07}: Exposure fusion weights (we use these)
	\item \cite{DBLP:journals/tog/TocciKTS11}: Hardware for HDR video using beamsplitters and multiple sensors
	\end{itemize}
}

In addition to fundamental video processing applications, our approach can also be used to create compelling, stylistic scene-space effects.

\paragraph{Inpainting and semi-transparency}
\label{sec:application:xray}
We can use our method to ``see-through'' objects by displaying content that is observed behind an object at another point in time.
This application requires that a user specifies which objects should be made transparent, either by providing per-frame binary image masks $\bm{M}$ where $\bm{M}(p)=1$ indicates that pixel should be removed, or a scene-space bounding region, and $\bm{M}(p)=0$ otherwise.
In the latter case, we project all samples that fall into the scene-space bounding region back into the original images to create $\bm{M}$.
We show in the supplemental video how one might create a scene-space mask with substantially less interaction than drawing 2D image masks, and compare our result to a state-of-the-art video inpainting method~\cite{DBLP:conf/eccv/GranadosKTKT12} using masks provided by the authors, Fig.~\ref{fig:inpainting}.

As we do not have a reference $s_{ref}$ in $\bm{S}(p)$ for the mask region, we instead compute an approximate reference sample by taking the mean of all samples,
\begin{equation}
s_{ref} = \frac{1}{|\bm{S}(p)|}\sum_{s\in\bm{S(p)}}{s}
\end{equation}
and weight samples with the following function,
\begin{equation}
w_{inpaint}(s) =  
\begin{cases}
\exp \left({-\frac{(s_{ref}-s)^2}{2\sigma^2}} \right) & \text{ when }  \bm{M}(s_p)=0 \\
0 & \text{ else } 
\end{cases}
\label{eq:inpainting}
\end{equation}
This computes a weighted combination of samples based on their proximity to the mean sample.
If we iterated this procedure, this would amount to a weighted mean-shift algorithm that converges on cluster centers in $\bm{S}(p)$. 
However we found that after two steps the result visually converges.
For inpainting, we use parameter values $\sigma_{rgb}=55$.
To achieve semi-transparent results, we add the standard multivariate weighting to the input frame $\bm{I}(p)$ and use $\sigma_{rgb}=80$, in order to emphasize similar color samples.

\begin{figure*}[t]
\setlength{\tabcolsep}{1px}
\centering
\def\myheight{3.8cm}
\begin{tabular}{*{4}{c@{\hspace{2px}}}}
{\begin{sideways}\parbox{\myheight}{\centering Computational Shutters}\end{sideways}} &
\includegraphics[height=\myheight,clip,trim=325 50 175 150]{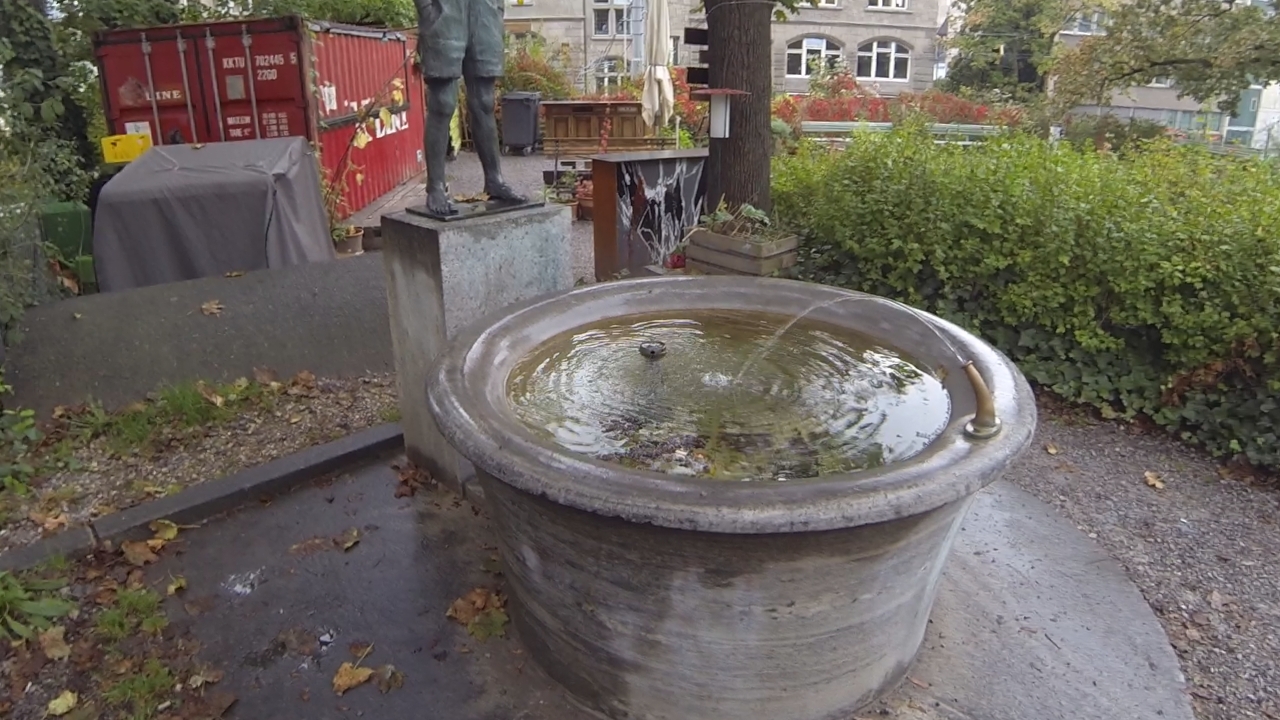} &
\includegraphics[height=\myheight,clip,trim=325 50 175 150]{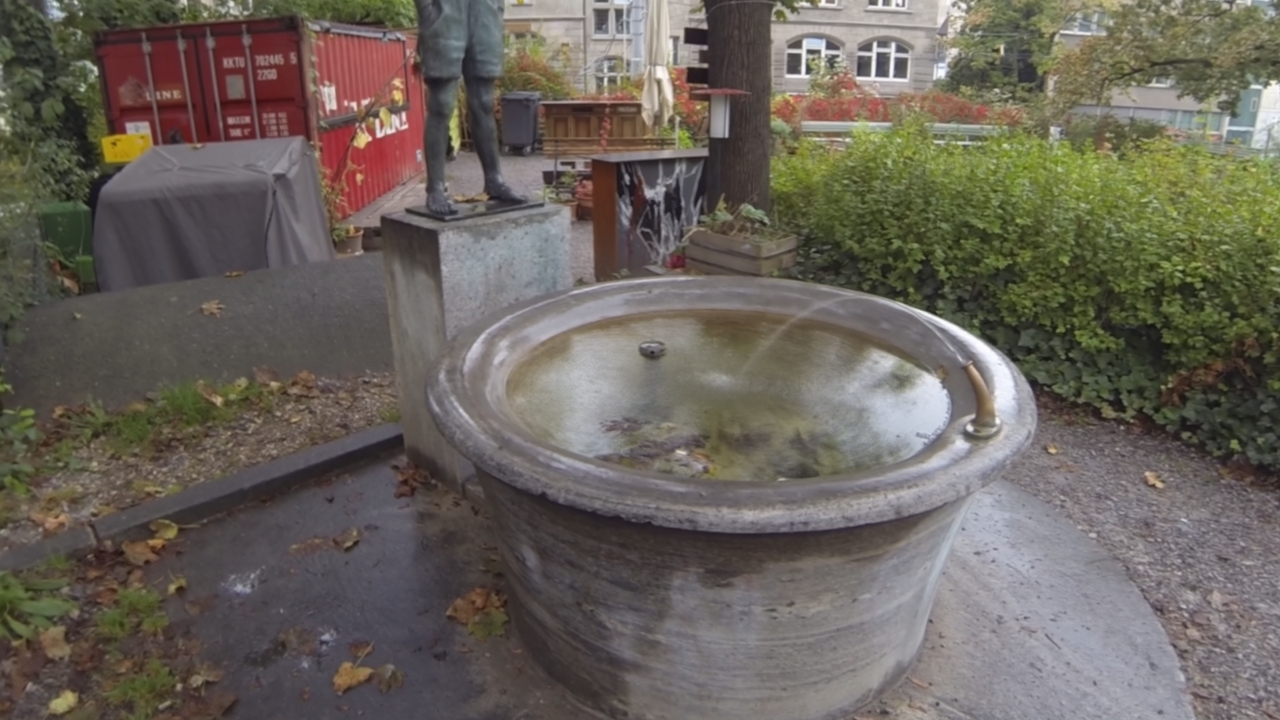} &
\includegraphics[height=\myheight,clip,trim=325 50 175 150]{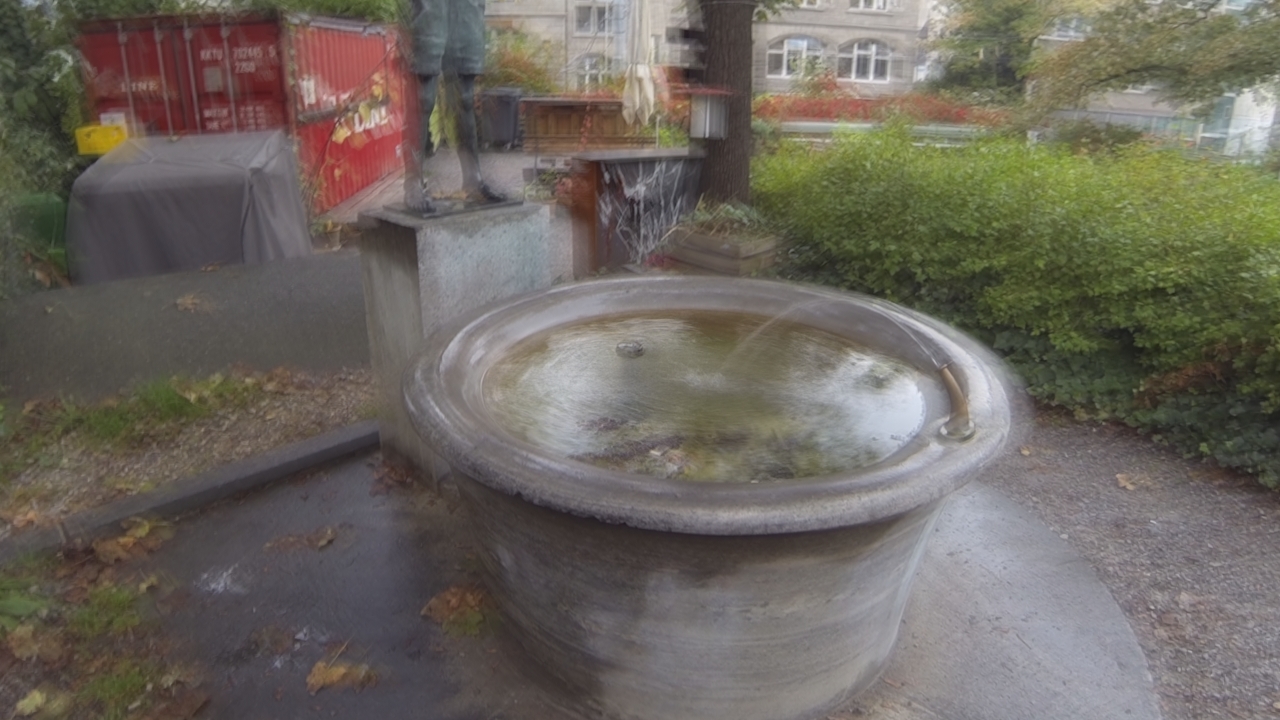} \\
& Input & Scene-space long exposure & Image-space long exposure \\
\end{tabular}

\begin{tabular}{*{4}{c@{\hspace{3px}}}}
{\begin{sideways}\parbox{\myheight}{\centering Action Shots}\end{sideways}} &
\includegraphics[clip,trim=200 0 80 0,height = \myheight]{images/compshutter/frame_00384_ours.jpg} &
\includegraphics[clip,trim=200 0 5 0,height = \myheight]{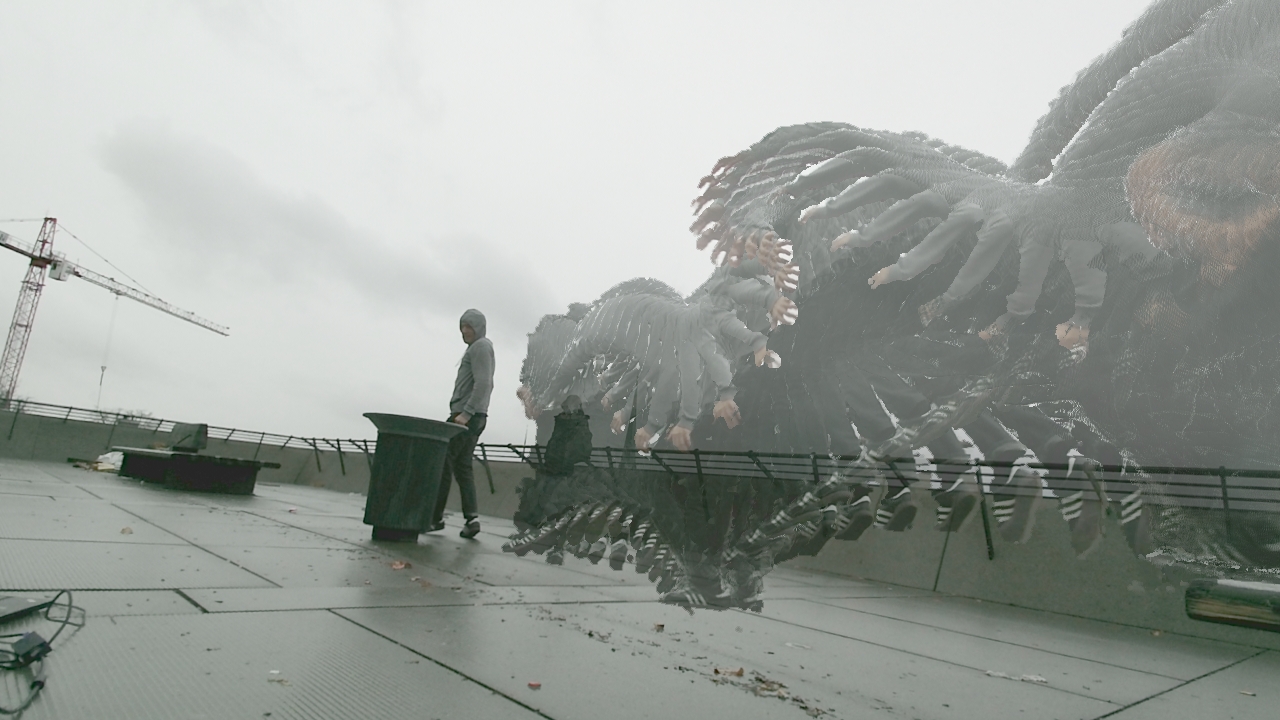} & 
\includegraphics[clip,trim=50 0 50 0,height = \myheight]{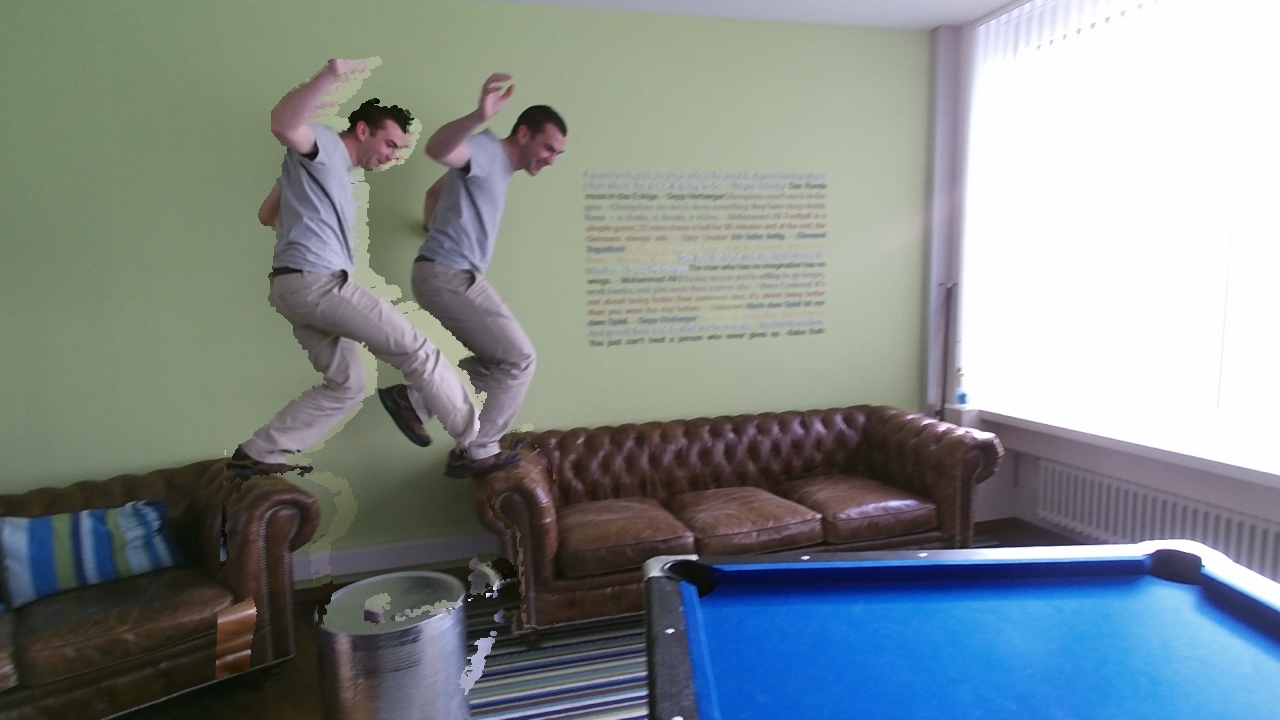} \\
\end{tabular}
	\caption{Computational shutter effects. In a \emph{scene-space} long exposure video, static objects remain sharp despite camera motion, while moving components like the fountain become blurred.  Image-space long exposure on the other hand, blurs the whole frame (top row).
	Scene-space processing enables effects such as action shots or motion trails, where the filtering step selects distinct frames in time  (bottom row). Because these effects live in scene-space, they behave correctly in terms of occlusions and perspective change.}
	\label{fig:computationalshutters}		
\end{figure*}

\paragraph{Computational scene-space shutters}
\label{sec:application:compshutter}
A ``computational shutter'' replaces the process of a camera integrating photons that arrive at a pixel sensor with a controlled post processing algorithm. 
By extending this concept into \emph{scene-space}, we can generate compelling results that are fully consistent over camera motion.
In this case, our weighting function is replaced by the shutter function that operates on the \emph{time} of each sample:
\begin{equation}
w_{compshutter}(s) =  \xi(s_f)
\label{eq:compshutter}
\end{equation}
where $\xi(s_f)$ is a box function in a typical camera. 
The most straightforward example is a scene-space long exposure shot, Fig.~\ref{fig:computationalshutters}. 
As opposed to image-space long exposure shots, time-varying components become blurred but the static parts of the scene remain sharp, even with a moving camera. 

\begin{figure}[t]
	\includegraphics[width=\linewidth]{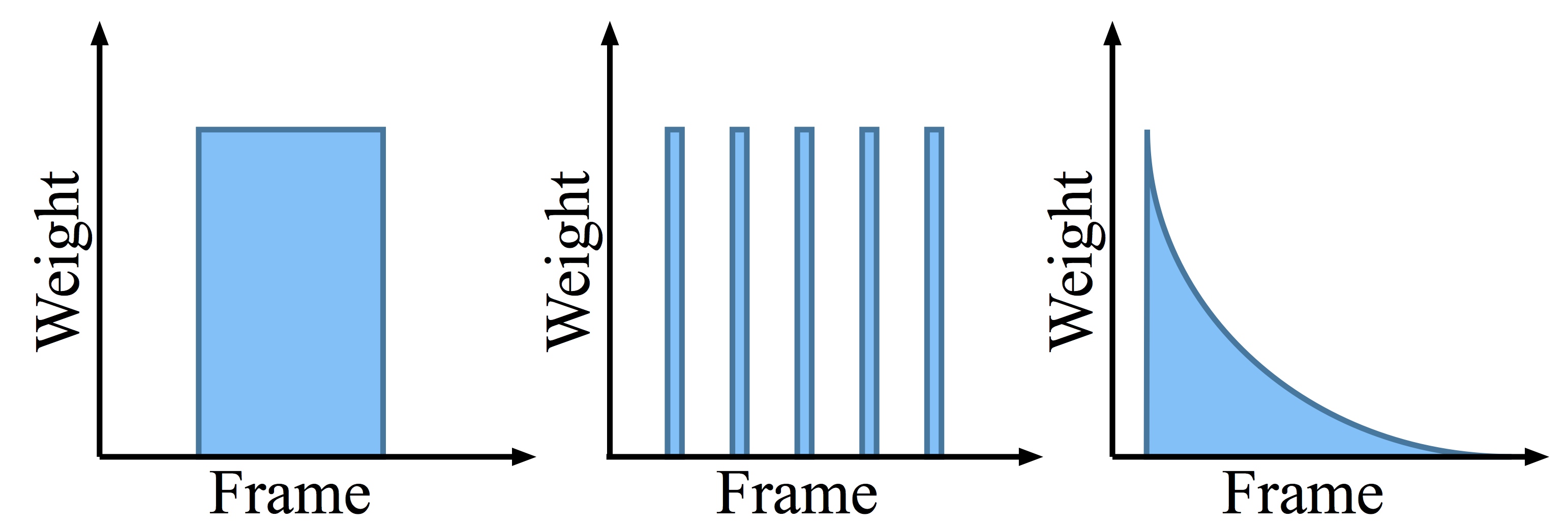}
\caption{\label{fig:shutterfunctions} A selection of ``computational shutter'' weighting functions, from left to right; a box filter equivalent to a regular camera shutter, an impulse train used to generate multiple exposure video action shots, and a long falloff used for motion trails.}
\end{figure}

We define possible alternatives for $\xi(s_f)$ visually in Fig.~\ref{fig:shutterfunctions}.
If we consider $\xi(s_f)$ to be an impulse train, and apply it only in a user-defined scene-space region, we can obtain ``action shot'' style videos. 
By using a long-tail decaying function, we can create trails of moving objects. 
These effects are related to video synopsis~\cite{DBLP:journals/pami/PritchRP08}, as they give an immediate impression of the motion of a scene.
In both cases, the temporally offset content behaves correctly with respect to occlusions and perspective changes. 
\final{When the computational shutter contains Dirac delta functions, this implies a \emph{projection} of content from one frame into another  (i.e., action shots in Fig.~\ref{fig:computationalshutters}). 
As only a single time instance is used, we cannot leverage repeated observations of these scene points, and we lose some of the robustness to noisy depth, leading to visible reprojection errors typical to depth image-based rendering. 
Nonetheless, this example highlights the expressiveness of our method by its ability to model complex scene-space effects using simple shutter functions.
For this, we require reasonable depth of moving foreground objects, which we acquire using a Kinect depth sensor, as it cannot be deduced from the video directly.
}

Until now, we have not explicitly addressed occlusions.
This is because inaccurate depth makes reasoning about occlusions difficult.
Instead we have relied on $s_{ref}$ and sample redundancy to prevent color bleeding artifacts.
However, using this approach for dynamic foreground objects, our method can only capture a single observation at a given moment of time.
Because we have neither a reference sample nor a significant number of samples with which to determine a reasonable prior, we use the following simple occlusion heuristic to prevent color bleed-through for scenes with reasonable depth values, e.g., from a Kinect.
We introduce the notion of a sample depth order $s_{ord}$, that is the number of samples in $\bm{S}(p)$ that are closer to $p$ than the current sample $s$,
\begin{equation}
s_{ord} =  \# \{q \in \bm{S}(p) \ \big\vert \ (p_{xyz}-q_{xyz})^2 < (p_{xyz}-s_{xyz})^2 \}.
\end{equation}
Our weighting function becomes:
\begin{equation}
w_{action} = \xi(s_f) \exp \left(-\frac{s_{ord}^2}{2\sigma_{ord}^2} \right).
\end{equation}
We use $\sigma_{ord}=10$.
This weighting function emphasizes the samples in the gathered frustum that are the closest to the output camera.

\ignore{	
	\paragraph{Motion trails}
	\label{application:motiontrails}
	
	While this work focuses on scene-space video processing, there are obvious applications of object space video processing.
	In order to do this, accurate scene flow is needed.
	This is still an open problem.
	\cite{DBLP:journals/tog/SchmidSBG10} motion trails on rendered content
	However, we experiment with a simple form of object space video processing, where we track a point in image-space and project that point into scene-space, for object space effects.
	In this, we show a synthetic ``light painting'' effect, where the motion of an object leaves a trail in scene-space, that is consistent in time as the camera moves.
	
	\paragraph{Scene-space consistency}
	\label{application:maskrefinement}
	\begin{itemize}
	\item rough masks can be refined based on scene-space voting
	\item inconsistent input can be regularized (see timelapse)
	\end{itemize}
}

\begin{figure}[t]
\setlength{\tabcolsep}{1px}
\centering
\def\myheight{3.8cm}
\newcommand{\vainclude}[1]{\includegraphics[width=\linewidth,trim = 0 30 0 110,clip]{#1}}
\begin{tabular}{*{1}{c@{\hspace{3px}}}}
\vainclude{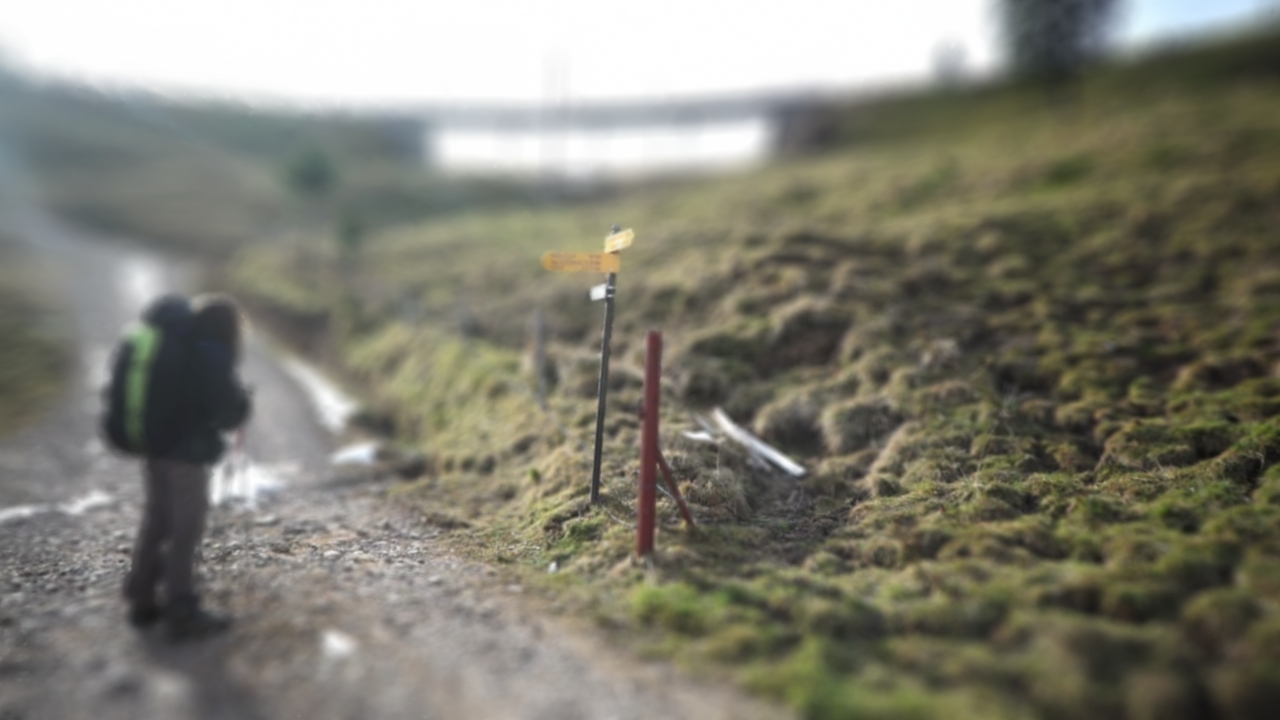}\\
\end{tabular}	
\caption{\label{fig:virtualaperture}Frame from a video with an added virtual aperture effect (narrow depth-of-field).}
\end{figure}

\paragraph{Virtual aperture}
\label{sec:application:aperture}
With appropriate weighting functions, we can also represent complex effects such as virtual apertures, exploiting the existence of our samples in a coherent scene-space.
To do this, we model an approximate physical aperture in scene-space and weight our samples accordingly, Fig.~\ref{fig:virtualaperture}. 
This allows us to create arbitrary aperture effects, such as focus pulls and focus manifolds defined in \emph{scene-space}.

\begin{figure}[t]
	\includegraphics[width=\linewidth]{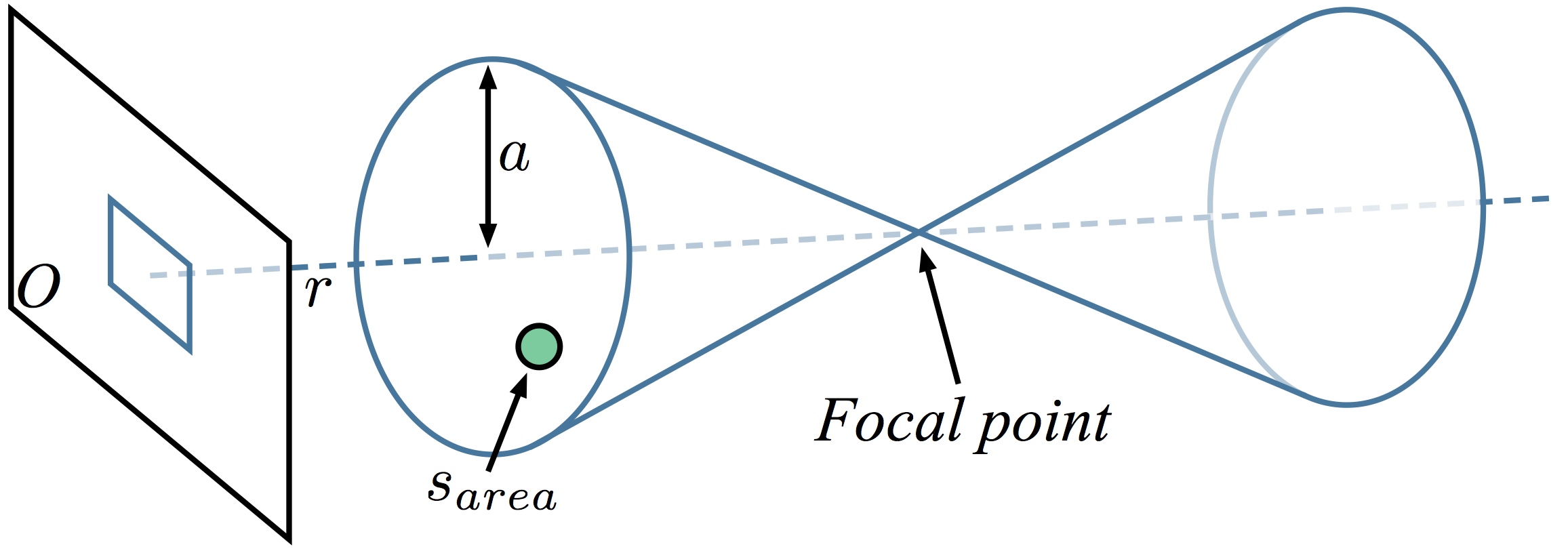}
	\caption{\label{fig:aperture} The shape of our virtual aperture weighting function is a double cone with the focal point at the center. The distance from this focal point along the viewing ray determines the size of the area from which samples are used.}
\end{figure}

We design a weighting function for an approximate virtual aperture as a double cone with its thinnest point $a_0$ at the focal point $z_0$, Fig.~\ref{fig:aperture}. The slope $a_s$ of the cone defines the size of the aperture as a function of distance from focal point;
\begin{equation}
a(z) = a_0 + |z_0-z| * a_s.
\end{equation}

To avoid aliasing artifacts we use the sample area $s_{area}$ introduced previously to weight each sample by the ratio of its size and the aperture size at its scene-space position. 
The intuition is that samples carry the most information at their observed scale.

With $r$ as the distance of $s_{xyz}$ along the camera viewing ray, and $q$ as the distance from the ray to $s$, we use 
\begin{equation}
w_{va} = \begin{cases}
 \frac{s_{area}}{\pi a(r)^2}  & \text{ when }  q < a(r)\\
0 & \text{ else }.
\end{cases}
\end{equation}
Many other formulations for synthetic apertures are conceivable and work on camera arrays demonstrated how it is possible to create compelling virtual aperture images and videos~\cite{DBLP:journals/tog/WilburnJVTABAHL05,vaish2005synthetic}.
In our case, we are not using multiple viewpoints at the same time instance, but rather samples captured from neighboring frames to compute aperture effects.

\ignore{

	\paragraph{Interactive control}
	We also experimented with an interface that would allow a user to ``paint'' on the video.
	These strokes are converted into positive or negative samples that either downweight or upweight nearby (in the 7D sense) samples.
	This provides an intuitive way to reduce ghosting (paint a positive sample on the frame you want), and to remove objects (paint a negative sample on an object to remove)
	\ow{how to discuss this?}
	This was used only on the datasets that are labeled as user interaction.

	\paragraph{Post Processing Optimization}
	\label{sec:application:postprocessing}
	\begin{figure}
	\missingfigure{}
	\caption{\label{fig:qpbo}Images showing dominant mode selection from QPBO}
	\end{figure}
	\begin{itemize}
	\item While we generate output samples one at a time, sometimes sharing information is desired
	\item we investigate one application, creating a MRF and solving it using QPBO (which is like graph cuts but works better on nonsubmodular energy)
	\item If our processing step, selects dominant modes of the sample sets, we can then perform some kind of image-space (or scene-space) optimization to select from these sample sets.
	\item We have tried this using QPBO, belief propagation and enforcing visual consistency.
	\item as an alternative approach, you can find modes along the depth values of a sample set
	\item these modes are noisy, but if you filter the modes, and use these in the weighting scheme, then you can get okay results
	\end{itemize}
}

%% file: implementation.tex
Our method was implemented in Python with CUDA bindings and tested on an Intel i7 3.2 GHz desktop computer, with 24GB RAM and a GeForce GTX 980 graphics card.
Datasets were captured using an iPhone 5s, GoPro Hero 3, Canon S100 point-and-shoot, and Sony $\alpha$7s camera.
In general, we process videos ranging from 200-1000 frames (6-30 seconds) long, all at 720p resolution.
Unless otherwise noted, all datasets were preprocessed with an automatic script that computes camera pose and data-term only depth maps.
Tab.~\ref{tab:time} gives the processing times for different steps of our method. 
\final{
The samples/pixel are averages observed in our experiments, and are determined by the required sampling regions; in inpainting we use a large temporal $\sigma_f$, accumulating samples over many frames, and with a virtual aperture, we gather from a bigger frustum, accumulating many samples over a large spatial region.		
The cost of gathering vs. filtering also depends on the application, but in most cases, we observed that roughly 80-95\% is spent gathering samples and the rest in filtering.
}
Despite the large amount of data considered, our prototype implementation shows that sampling based scene-space approaches can generate results in a \emph{reasonable} amount of time. 
Further optimizations are possible for performance critical applications.

%% file: evaluation.tex
We present a general framework for implementing a variety of effects efficiently and intuitively. %
We do not claim that our approach always performs better than any of the application specific state-of-the-art methods that we compare to, only that we can generate compelling video processing effects.
For this reason, we provide videos in the supplemental material as validation of our approach.

\begin{figure*}[t]
\setlength{\tabcolsep}{1px}
\centering
\newcommand{\robustinclude}[1]{\includegraphics[height=3.2cm,trim = 400 175 740 350,clip]{#1}}
\newcommand{\brobustinclude}[1]{\includegraphics[height=3.2cm,trim = 400 175 680 350,clip]{#1}}
\begin{tabular}{*{6}{c@{\hspace{3px}}}}
\robustinclude{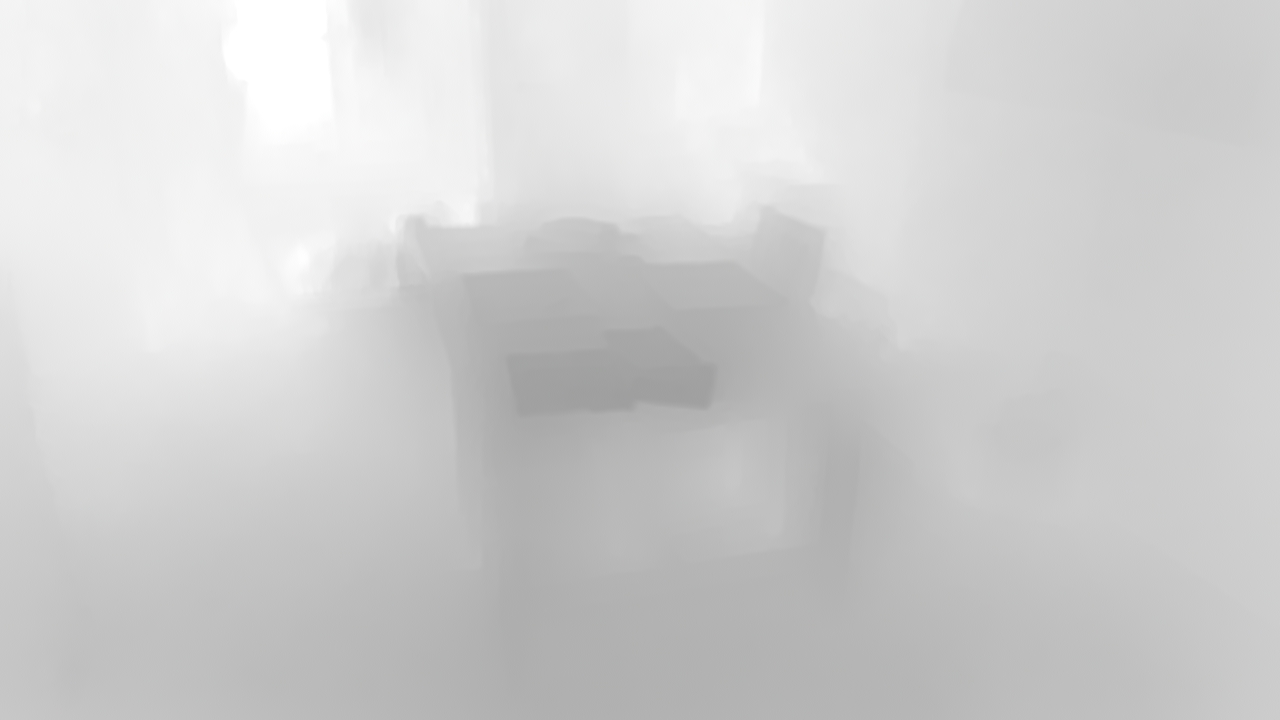} &
\robustinclude{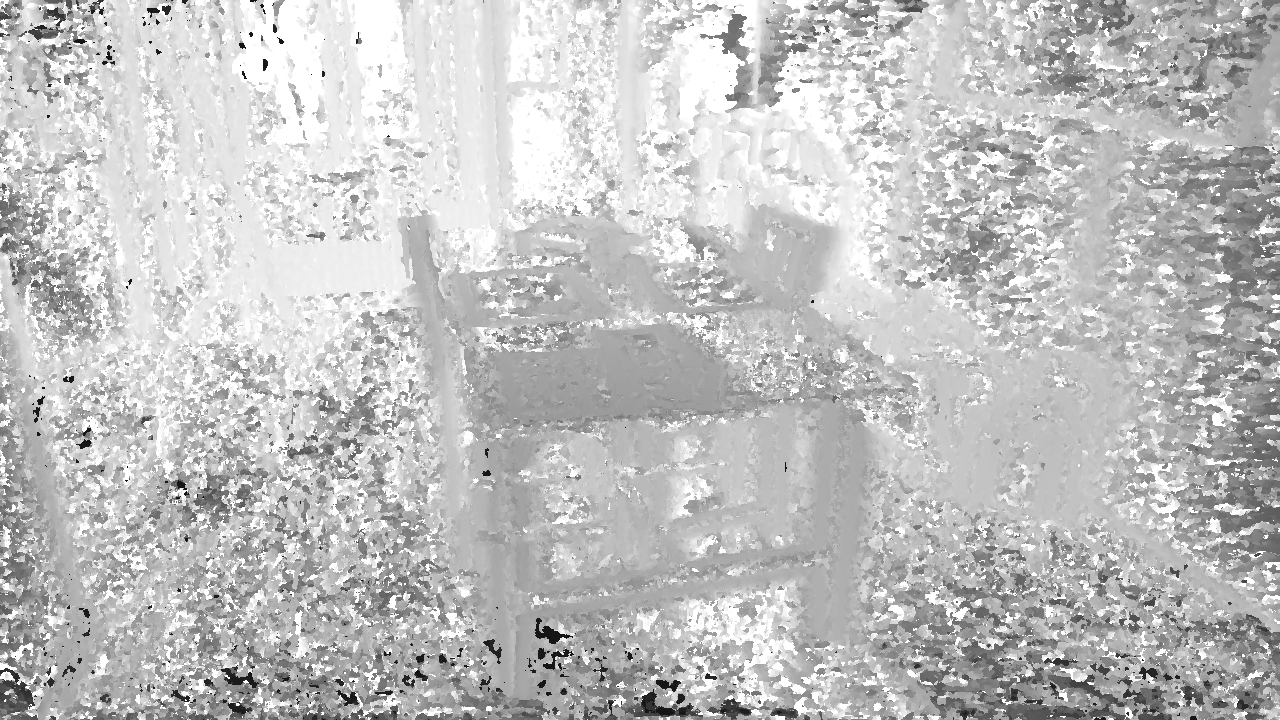} &
\robustinclude{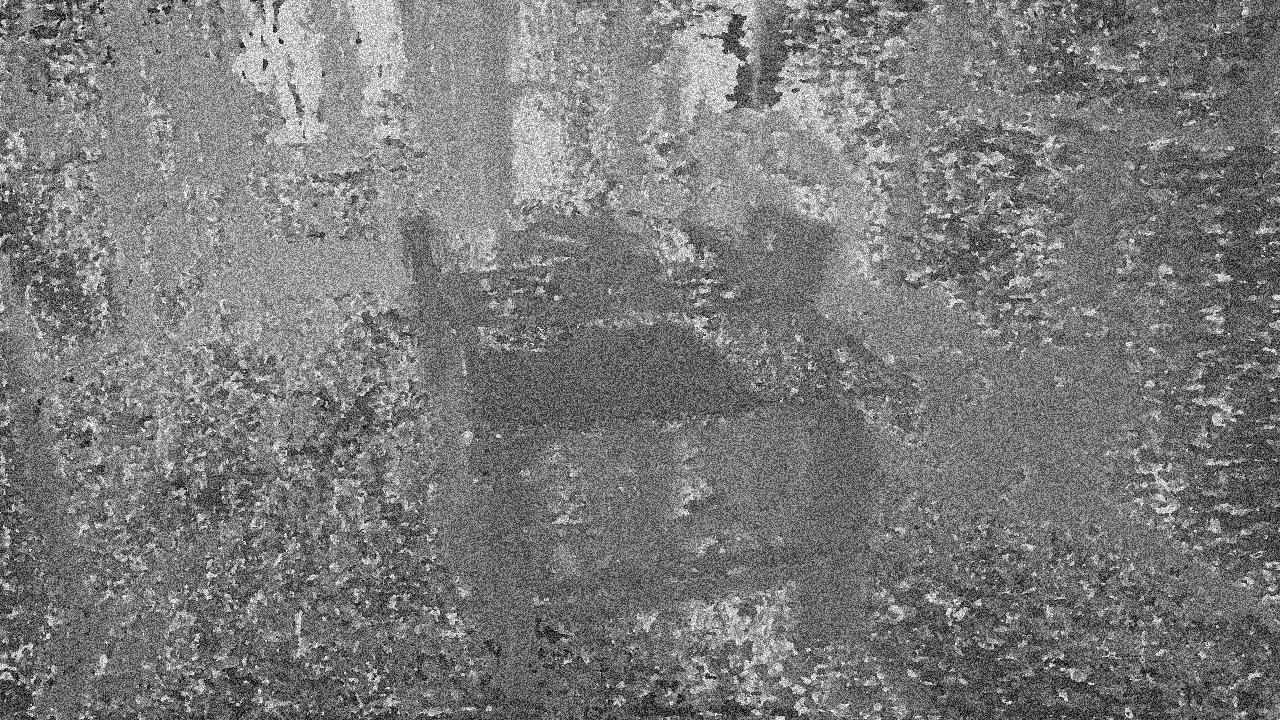} &
\brobustinclude{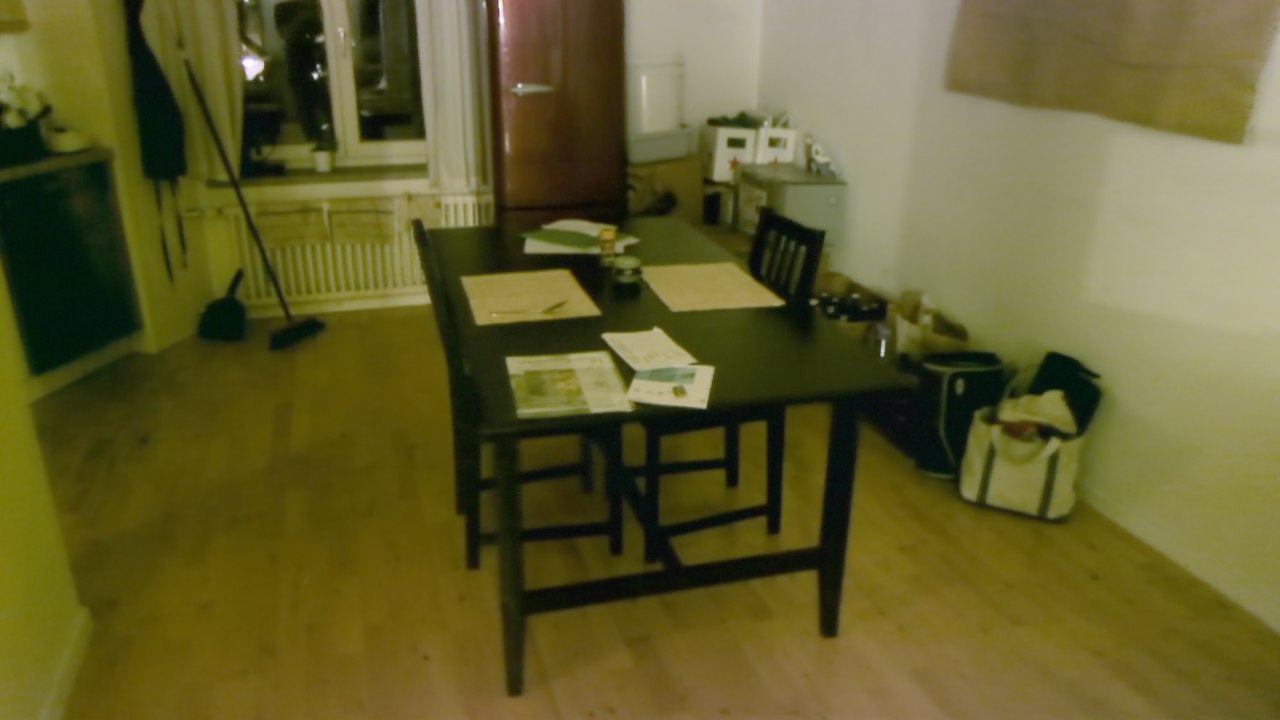} &
\brobustinclude{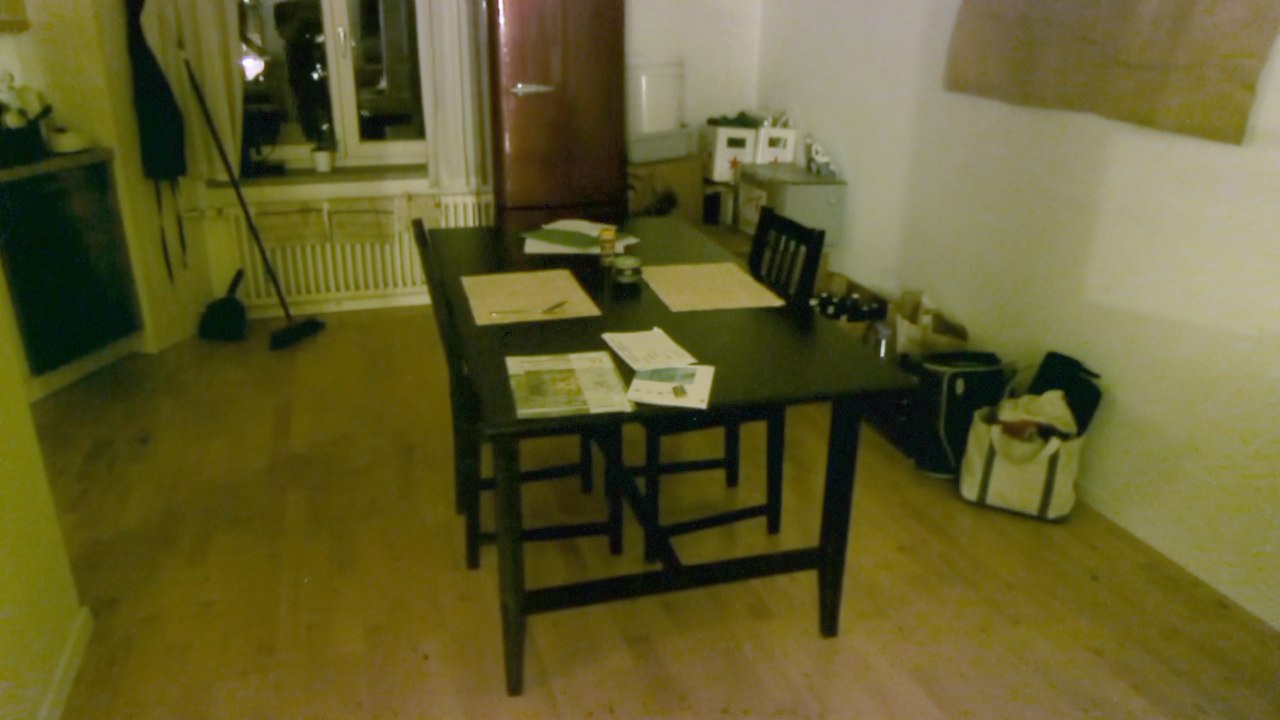} &
\brobustinclude{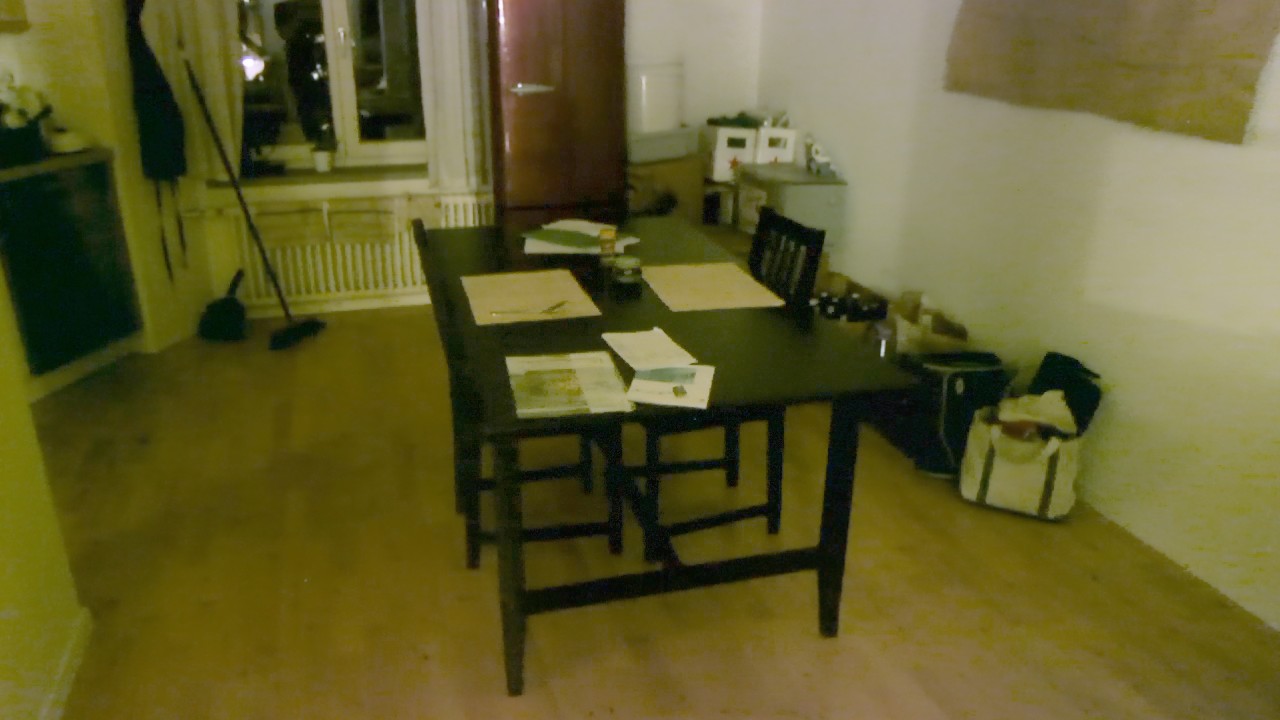} \\
(a) & (b) & (c) & Denoised with (a) & Denoised with (b) & Denoised with (c) \\
\end{tabular}
\caption{\label{fig:robustness}
\final{Evaluation of denoising with varying quality depth maps: globally smooth (computed using Nuke) (a), data-term only depth (b), and (b) with added high frequency noise (c). While the denoised output is similar in many places, some artifacts are visible when zoomed in. For example, when denoised with the globally smoothed (a) or added noise (c) results, more of the background color bleeds in from behind the table legs, and in (c) the final result has substantially more high frequency noise.}
}
\end{figure*}

\begin{figure*}[t]
\setlength{\tabcolsep}{1px}
\centering
\newcommand{\ameaninclude}[1]{\includegraphics[height=3.5cm,trim = 0 100 650 100,clip]{#1}}
\newcommand{\bmeaninclude}[1]{\includegraphics[height=3.5cm,trim = 850 300 0 100,clip]{#1}}
\begin{tabular}{*{4}{c@{\hspace{3px}}}}
\ameaninclude{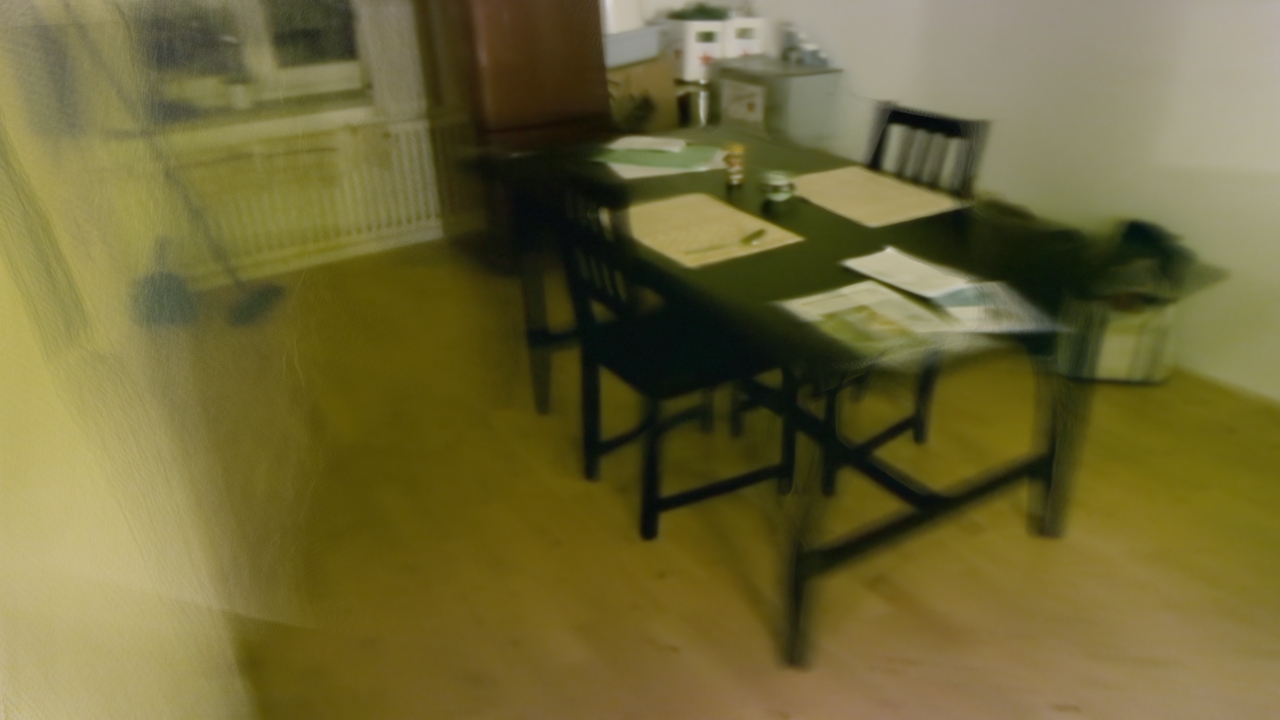} &
\ameaninclude{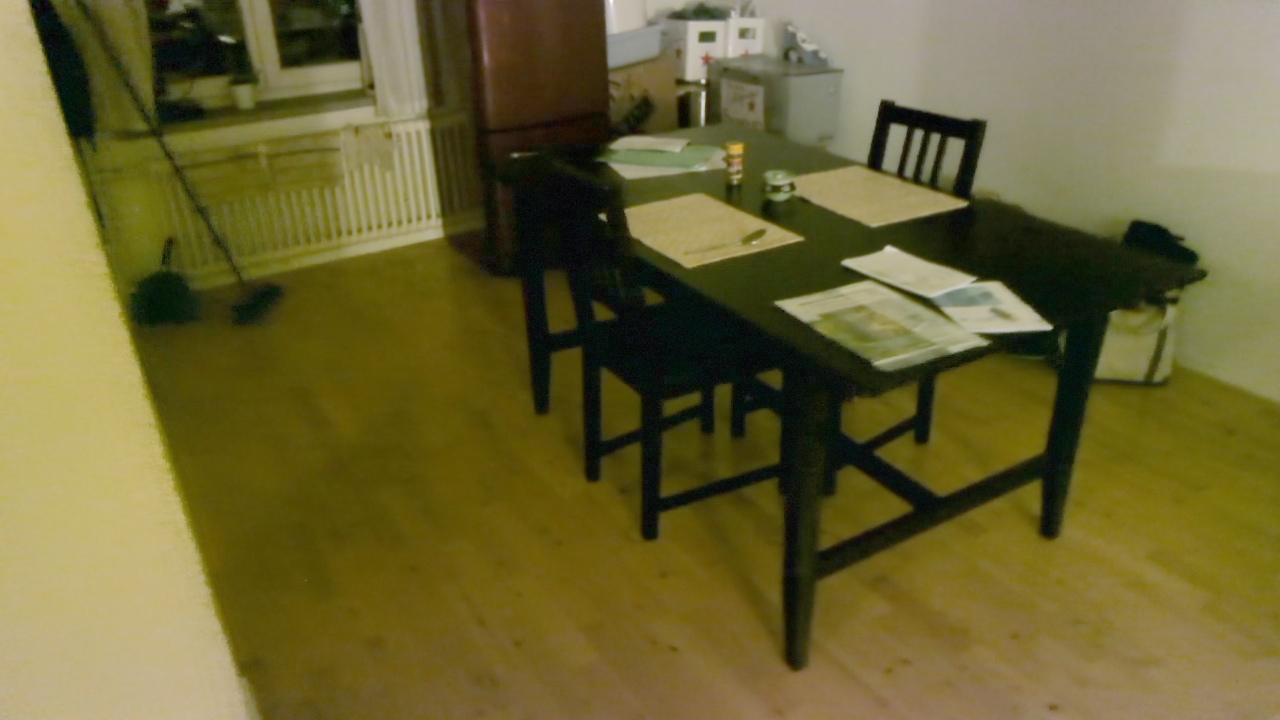} &
\bmeaninclude{images/denoising/frame_0110_mean.jpg} &
\bmeaninclude{images/denoising/frame_0110_gauss.jpg} \\
a) Mean of all samples & Our method & b) Mean of all samples & Our method \\
\end{tabular}
\caption{\label{fig:mean} Averaging the color of all gathered samples (mean) generates poor quality results because it includes many incorrect samples due to 3D projection errors. Using our filtering, the contribution of these samples can be greatly reduced.}
\end{figure*}

\setlength{\tabcolsep}{1.2mm}
\begin{table}[t]
\centering
\begin{tabular}{lrrr}
\toprule
 & samples/pixel & sec./frame \\
\midrule
Preprocessing & - & 1.5 \\
Depth Computation & - & 28.5 \\
\midrule
Denoising & 500 & 3.4 \\
Deblurring & 250 & 8.9 \\
Action shots & 100 & 4.7 \\
Video Inpainting & 1000 & 16.0 \\
Virtual Aperture & 12000 & 10.2 \\
Motion Trails & 600 & 29.0 \\
Super resolution ($9\times$resolution) & 800 & 140.1 \\
\bottomrule
\end{tabular}
\caption{Processing time for different applications of our method. 	
	Preprocessing is the total time spent on automatic extraction of frames, lens undistortion, and camera pose estimation.	
	Runtimes are dependent on scene structure, camera motion, sampling parameters, so timings are only given as an estimate of the runtime the algorithm can be expected to perform at.
	}
\label{tab:time}
\end{table}

We additionally analyzed the performance of our method over a range of different sources of 3D information. 
We experimented with alternate camera pose estimation tools (VisualSFM, PFTrack) with similar results.
\final{
In Fig.~\ref{fig:robustness}, we show a denoising result using a variety of depth maps, from over-smoothed to highly noisy. 
This example shows that our method degrades gracefully as the quality of depth is reduced, leading to only subtle artifacts visible as residual noise and color bleeding from occluded scene objects.
}
We further demonstrate the effect of our robust filtering step in Fig.~\ref{fig:mean}, by comparing our filtering to averaging all gathered samples.
With a simple mean, errors in scene-space information are clearly visible. 

In most applications, we use $5<\sigma_{f}<20$, which implies that we are only using information up to about 60 frames on either side of the output frame. 
This limitation is due to drift in the camera calibration. 
After more frames, we found that the reprojection errors become worse and worse and gathering these samples did not help much. 
By choosing lower $\sigma_f$, our approach can be robust to this kind of drift, at the expense of having fewer samples available for filtering.
However, given drift-free calibrations, e.g., from pose estimation techniques employing loop closure, we could take advantage of additional frames.

\subsection{Limitations and future work}

\begin{figure}[t]
\centering
\def\myheight{3cm}
\includegraphics[width=.49\linewidth,clip,trim=750 450 200 100]{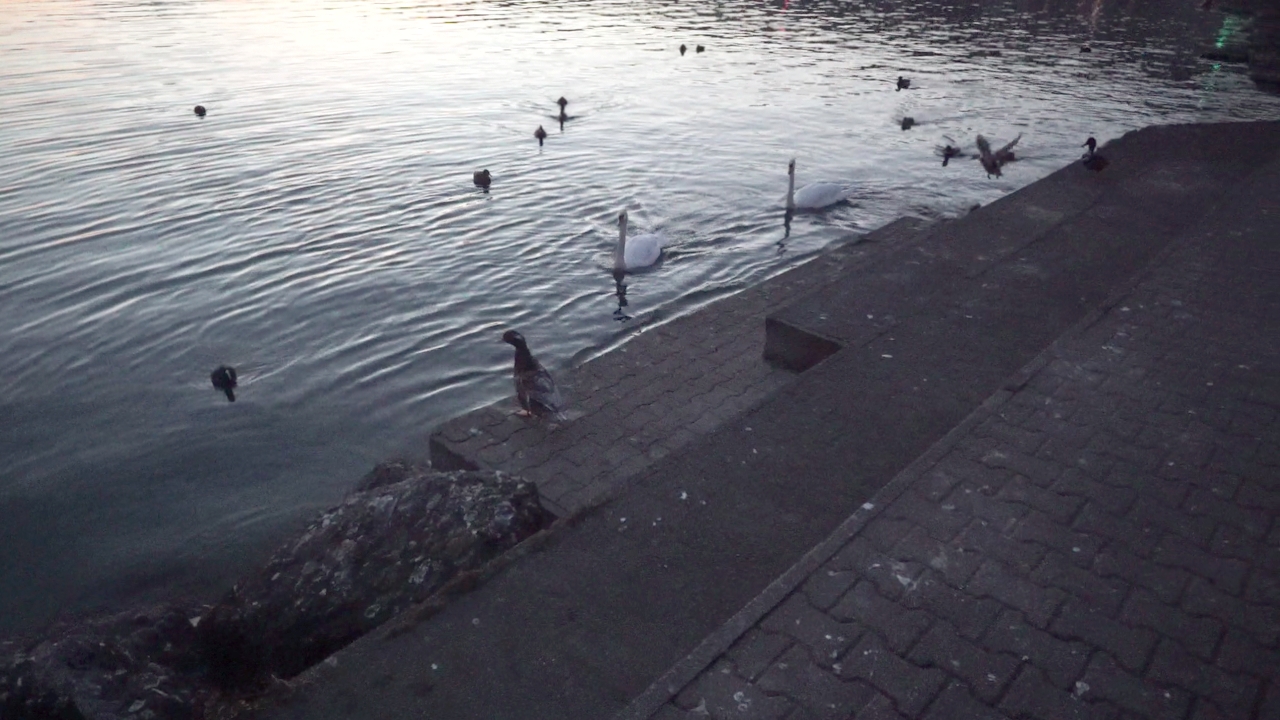}
\includegraphics[width=.49\linewidth,clip,trim=750 450 200 100]{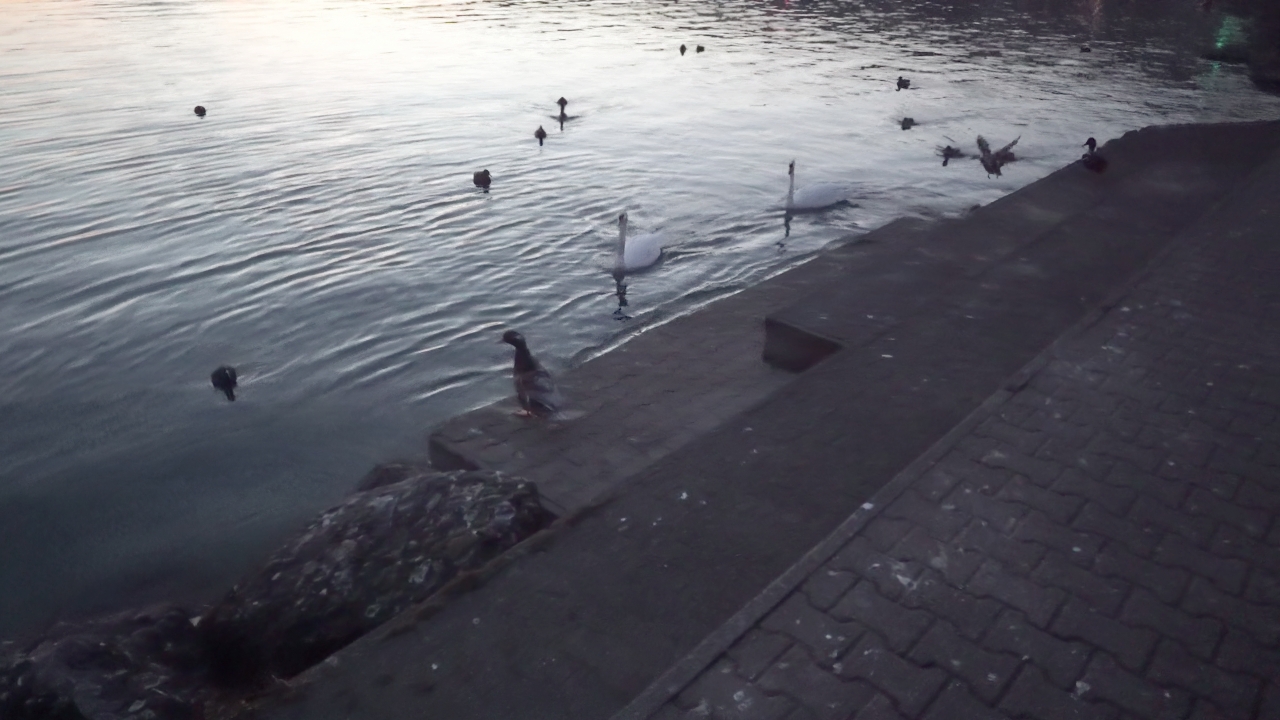}\\
\includegraphics[height=\myheight,clip,trim=400 850 2000 400]{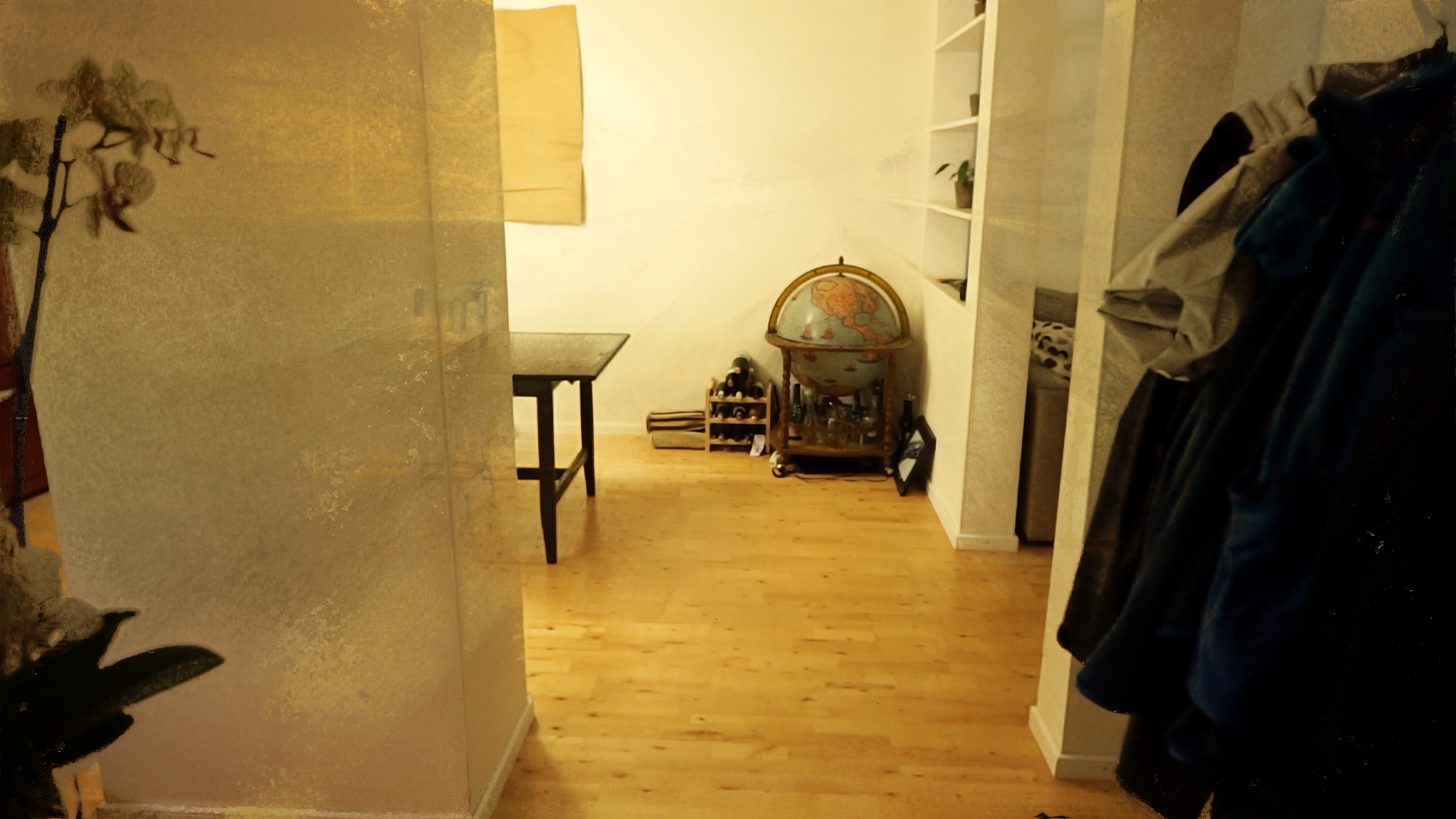}
\includegraphics[height=\myheight,clip,trim=290 00 550 350]{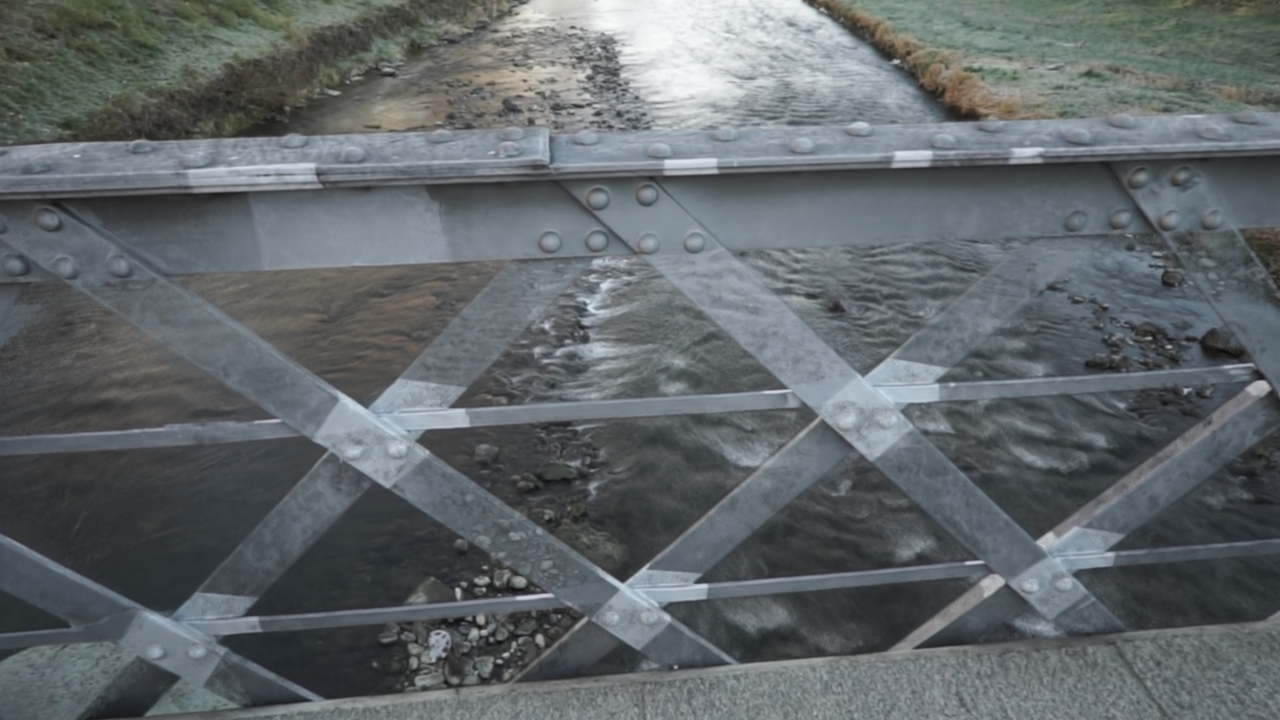}
\caption{\label{fig:lim}The top row shows an example of a rapidly moving object.  Our method cannot denoise the flapping bird as excessive motion prevents us from gathering reliable samples from other frames.  Despite this, we can denoise the rest of the scene and avoid ghosting of the bird. Occlusions can be an additional problem, as we gather samples both on and behind objects. We can reduce their contribution using depth information, or the reference sample.}
\end{figure}

Our approach has several limitations, and numerous directions for promising future work.
Our framework assumes that the color of scene points stays constant over a local temporal window, i.e., scene changes due to lighting variation or object motion are not explicitly modeled.
Despite this, the robust nature of the filtering step enables us to generate realistic-looking results even when these assumptions are violated. 
\final{
However, for fast moving objects, our ability to collect samples is limited to the current frame only, and as a result our filtering step can at best recreate the original frame, Fig.~\ref{fig:lim}.
For the action shots example, samples are gathered from a single, distant frame and no valid reference samples exist, which can lead to standard reprojection errors, clearly visible in Fig.~\ref{fig:computationalshutters}.
}

One solution would be to gather samples in \emph{object space} rather than scene-space.
Doing so however, requires computing accurate dense scene flow (scene-space optical flow).
Computing scene flow is a challenging problem, and while there has been exciting recent work showing high quality 3D trajectories~\cite{joo2014map}, these are often achieved with specialized acquisition scenarios (in this case a 480 camera dome). 
Nonetheless, as scene flow becomes more available, our approach is well suited to take advantage of it; by integrating the motion of the scene into the sample gathering step we can support fully dynamic scenes as well.

Occlusions are another inherent difficulty when working with unstructured point clouds.
As we gather samples throughout the entire output pixel frustum, we collect occluded samples that project \emph{behind} objects if these regions become visible at another time in the video.
If a good reference sample $s_{ref}$ is not available, these can show up as color bleeding in the foreground, Fig.~\ref{fig:lim}.
With sufficiently accurate depth information however, (e.g, in the action shots application), the contribution of these occluded samples can be reduced, allowing us to prevent unwanted color bleeding. 

\begin{figure}
\centering
\def\myheight{3cm}
\includegraphics[width=.49\linewidth,clip,trim=100 300 800 250]{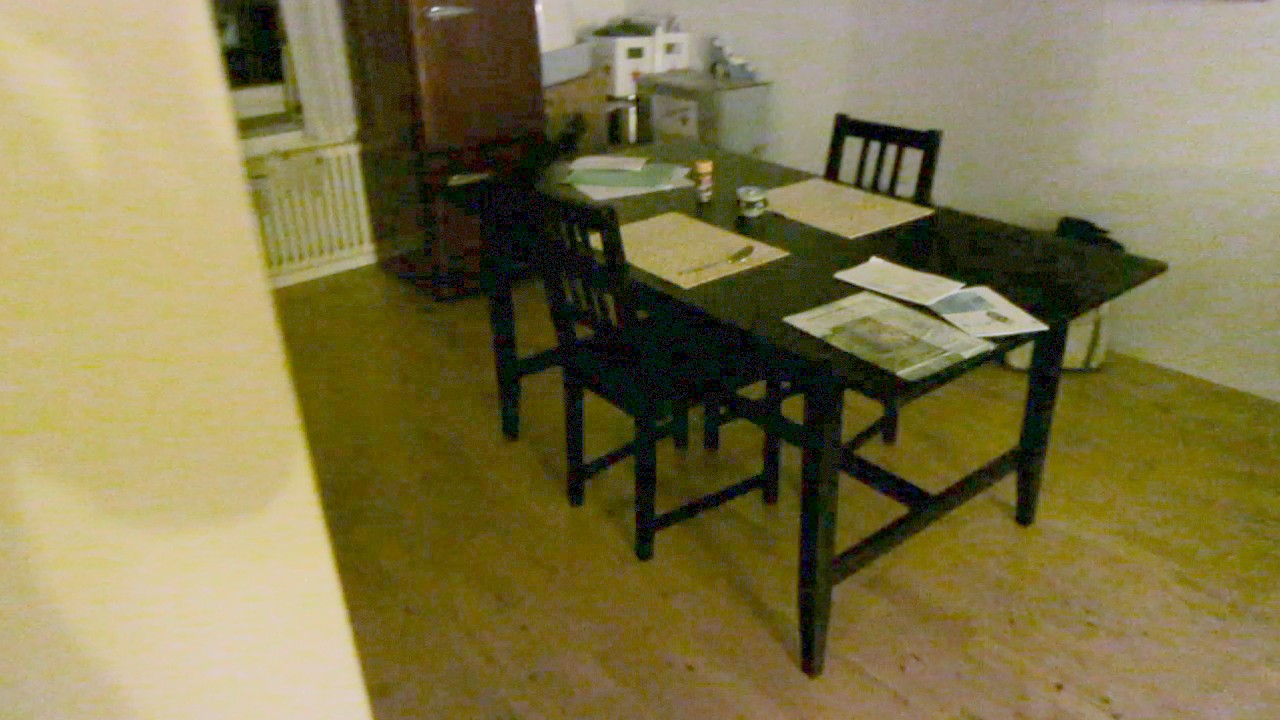}
\includegraphics[width=.49\linewidth,clip,trim=100 300 800 250]{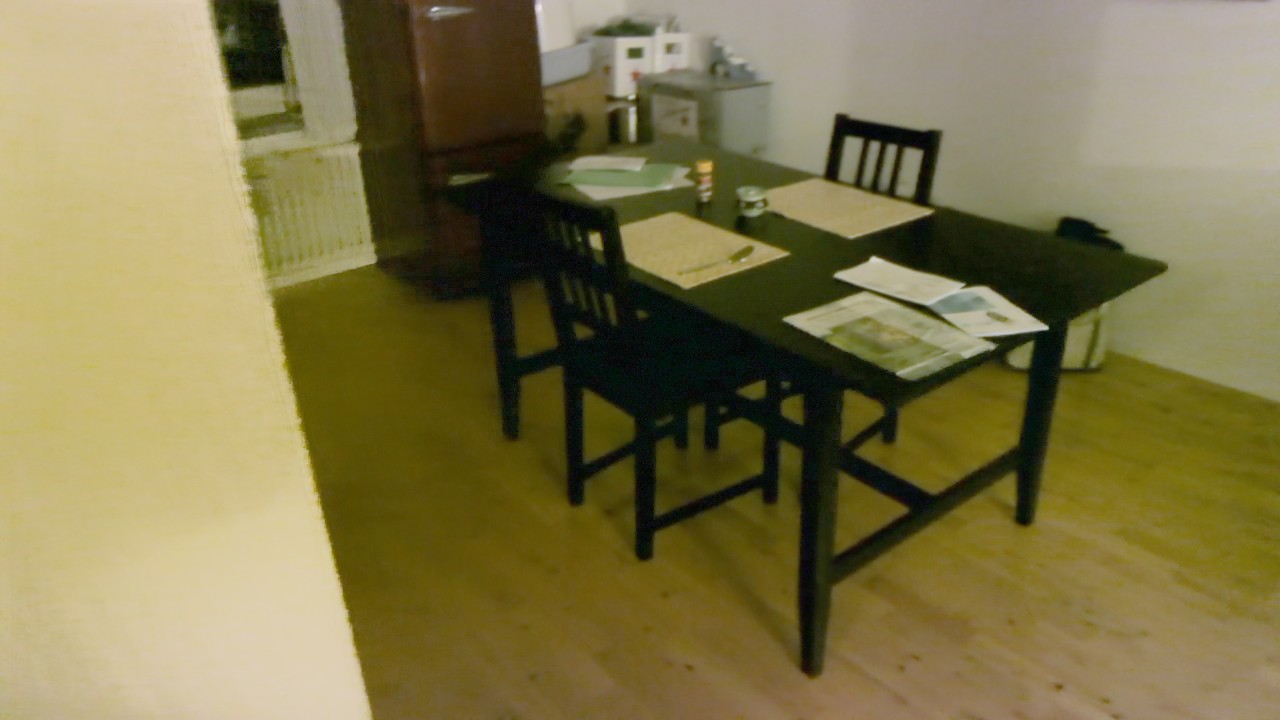}\\
\caption{\label{fig:limitation2}\final{The blurred edge of the wall in the noisy input image (left), makes both depth estimation and the reference sample unreliable, which can cause high frequency artifacts (right).}}
\end{figure}

Our approach computes neighboring output pixel colors independently.
This is important for efficient parallelization, but means that it is not straightforward to enforce higher-level image-space consistency in the final result.
\final{
This can lead to high frequency artifacts, for example in Fig.~\ref{fig:inpainting} and the \textsc{Horse} sequence in the supplemental material, where speckling is visible in regions with high contrast between foreground and background colors.
In cases where both the depth and reference sample color are incorrect, object boundaries can become distorted as the weighting of sample sets varies from pixel to pixel.
This effect is visible in Fig.~\ref{fig:limitation2} and in the \textsc{Kitchen} sequence in the supplemental video, where the edge of the wall becomes distorted.
}
We can mitigate this problem to some extent by controlling the size of the pixel frustums; by having overlapping frustums between neighboring pixels, neighboring sample sets will share samples, achieving a limited amount of spatial consistency. 
However, explicit methods that solve for image-space consistency, e.g., by solving for pixel colors as a MRF considering neighboring output pixels as being connected, would be an interesting area for future work.

Finally, our method requires scene-space information, which can be seen as an additional burden.
In fact, for some videos, existing approaches are not sufficiently robust to determine camera pose and depth values automatically.
As these technologies improve, however, more and more scenes will be suitable for automatic sampling-based processing.
In professional environments, these problems are solved on a daily basis with a high degree of accuracy using mature tools and skilled operators, and our approach can directly be integrated into such production pipelines.

%% file: conclusion.tex
We have presented a general framework that allows for robust scene-space video processing in the presence of inaccurate 3D information.
Our method is based on simple, transparent algorithms that can take advantage of the large volumes of data that comes with HD videos.
We have demonstrated the generality of our approach for different fundamental video processing applications and compared the results to state-of-the-art methods specifically designed to the respective tasks.
Additionally, we have shown several advanced video processing operations such as video inpainting, action shots, and virtual apertures. 
Each of these applications was computed on real world, hand-held footage captured with consumer-grade cameras, demonstrating the robustness of our approach and general applicability both to mass market and professional requirements.
We believe that our novel scene-space processing approach will enable new video applications that were previously impossible, limited or could not be fully explored because of inevitably unreliable depth information.
\final{Video results and datasets are available on the project website to facilitate future research.}